\pdfoutput=1
\documentclass[11pt]{article}

\usepackage[preprint]{acl}

\usepackage{times}
\usepackage{latexsym}
\usepackage{amsmath}
\usepackage{listings}
\usepackage[T1]{fontenc}
\usepackage[utf8]{inputenc}
\usepackage{booktabs}
\usepackage{cleveref}
\usepackage{graphicx}
\usepackage{multirow}
\usepackage{tablefootnote}
\usepackage{caption}
\usepackage{subcaption}
\usepackage{url}
\usepackage{makecell}
\usepackage{color,colortbl}
\usepackage{enumitem}
\newcommand\blfootnote[1]{%
  \begingroup
  \renewcommand\thefootnote{}\footnote{#1}%
  \addtocounter{footnote}{-1}%
  \endgroup
}
\usepackage{hyperref}

\usepackage{microtype}

\usepackage{inconsolata}

\title{StableToolBench: Towards Stable Large-Scale Benchmarking on \\
Tool Learning of Large Language Models}

\author{
Zhicheng Guo$\textsuperscript{\rm $1,2$}$,
Sijie Cheng$\textsuperscript{\rm $1,2,3$}$, 
Hao Wang$\textsuperscript{\rm $4$}$, Shihao Liang$\textsuperscript{\rm $5$}$, Yujia Qin$\textsuperscript{\rm $1$}$, \\
\textbf{Peng Li}$\textsuperscript{\rm $2$}$,
\textbf{Zhiyuan Liu}$\textsuperscript{\rm $1$}$,
\textbf{Maosong Sun}$\textsuperscript{\rm $1$}$,
\textbf{Yang Liu}$^{1,2,6}$
\\
 $\textsuperscript{\rm $1$}$Dept. of Comp. Sci. \& Tech., Institute for AI, Tsinghua University, Beijing, China\\
 $\textsuperscript{\rm $2$}$Institute for AI Industry Research (AIR), Tsinghua University, Beijing, China \\
 $\textsuperscript{\rm $3$}$01.AI  $\textsuperscript{\rm $4$}$Google
 $\textsuperscript{\rm $5$}$The University of Hong Kong \\
  $\textsuperscript{\rm $6$}$Jiangsu Collaborative Innovation Center for Language Competence, Jiangsu, China \\
 \{\texttt{guo-zc21}, \texttt{csj23}\}\texttt{@mails.tsinghua.edu.cn}
}

\begin{document}
\maketitle
\begin{abstract}
% This document is a supplement to the general instructions for *ACL authors. It contains instructions for using the \LaTeX{} style files for ACL conferences. 
% The document itself conforms to its own specifications, and is therefore an example of what your manuscript should look like.
% These instructions should be used both for papers submitted for review and for final versions of accepted papers.
% This document is a supplement to the general instructions for *ACL authors. It contains instructions for using the \LaTeX{} style files for ACL conferences. 
% The document itself conforms to its own specifications, and is therefore an example of what your manuscript should look like.
% These instructions should be used both for papers submitted for review and for final versions of accepted papers.
% This document is a supplement to the general instructions for *ACL authors. It contains instructions for using the \LaTeX{} style files for ACL conferences.
Large Language Models (LLMs) have witnessed remarkable advancements in recent years, prompting the exploration of tool learning, which integrates LLMs with external tools to address diverse real-world challenges. Assessing the capability of LLMs to utilise tools necessitates large-scale and stable benchmarks. However, previous works relied on either hand-crafted online tools with limited scale, or large-scale real online APIs suffering from instability of API status. To address this problem, we introduce StableToolBench, a benchmark evolving from ToolBench, proposing a virtual API server and stable evaluation system. The virtual API server contains a caching system and API simulators which are complementary to alleviate the change in API status. Meanwhile, the stable evaluation system designs solvable pass and win rates using GPT-4 as the automatic evaluator to eliminate the randomness during evaluation. Experimental results demonstrate the stability of StableToolBench, and further discuss the effectiveness of API simulators, the caching system, and the evaluator system. 
\end{abstract}

\section{Introduction}
% Framework:
% \begin{itemize}
%     \item Tool learning empowers LLM.
%     \item However, how to benchmark LLMs on tool usage remains an open issue. There exists a trade-off between questioning reality (To Be Defined/Changed) and the number of APIs but real-life tool using needs both. Previous works fall on one side.
%     \item We propose a new benchmark that strikes a balance between them. including scenario-based question construction, cache-based large-scale API pool and query satisfaction-oriented evaluation. We also propose a new strong baseline....
% \end{itemize}
% \textbf{\textcolor{red}{Framework:}}

% Large Language Models (LLMs, \citealp{brown2020language, geminiteam2023gemini,openai2023gpt4, touvron2023llama}) have demonstrated considerable achievements across a diverse array of tasks, such as commonsense reasoning~\cite{weng2023large, ling2023deductive} and coding~\cite{chen2021evaluating, rozière2023code}. When augmented with auxiliary tools including online search engines and external computational models, LLMs demonstrate enhanced performance in more complex tasks~\cite{nakano2022webgpt, yao2023react, lu2023chameleon}. Therefore, at the time when tool learning becomes more powerful, benchmarking LLMs in the context of tool learning has been increasingly important.

With the rapid developments of Large Language Models (LLMs; \citealp{brown2020language, geminiteam2023gemini,openai2023gpt4, touvron2023llama}), tool learning which leverage LLMs to schedule a variety of external tools has attracted enormous attention~\cite{nakano2022webgpt, yao2023react, lu2023chameleon}.
\blfootnote{Project: 
\href{https://zhichengg.github.io/stb.github.io/}{\texttt{zhichengg.github.io/stb.github.io/}}}\blfootnote{GitHub: \href{https://github.com/THUNLP-MT/StableToolBench}{\texttt{THUNLP-MT/StableToolBench}}} 
Previous studies~\citep{hao2023toolkengpt, hsieh2023tool, schick2023toolformer, tang2023toolalpaca} aim to augment LLMs with tools to enhance performance on conventional natural language processing (NLP) downstream tasks, while recent work~\citep{qin2023webcpm, NEURIPS2022_82ad13ec_webshop, cai2024large} primarily focus on solving real-world scenarios that require the use of tools.
In general, tool learning complements the capabilities of vanilla LLMs and bridges the gap to real-world applications.

To assess the capability of LLMs to use tools, a series of tool learning benchmarks have been introduced. 
Several pioneering studies have heavily relied on human-crafted offline tools~\cite{gpt4tools, xu2023tool} or hand-selected online tools~\cite{li2023api, li2023apibank, chen2023teval}.
While these tools are high-quality, their scale remains relatively small, thereby limiting their ability to accurately reflect real-world scenarios.
To address this limitation, subsequent studies~\citep{tang2023toolalpaca, ye2024tooleyes, qin2023tool} have advocated for leveraging extensive collections of online tools that span across various domains.
Owing to the increased scale, the automatic evaluation of tool learning has moved closer to real-world scenarios.
However, concerns have been raised regarding the stability of these online tools, which has implications for the reproducibility and comparability of benchmark performance over time\footnote{According to its \href{https://www.oed.com/dictionary/benchmark_n?tab=meaning_and_use}{definition}, benchmarks should remain stable, and the model performance assessed on them must be comparable over time.}. 
For instance, the well-recognised ToolBench\footnote{We use  ToolEval2 in ToolBench as the benchmark.}~\cite{qin2023toolllm} has shown performance discrepancies that cannot be reproduced months after its release, as analysed in ~\Cref{sec:pre_analysis_performance}.
This is even more important when faced with a complex environment, where APIs and tools keep changing while the evaluation should maintain its consistency across time.

Existing large-scale benchmarks may struggle to provide stable evaluations for various reasons. We propose several hypotheses for this issue.
Firstly, the complexity of tasks involving tool usage makes it challenging for the common automatic evaluator, \texttt{gpt-3.5}, to function effectively as a discriminator. As discussed in~\Cref{sta_eval}, the evaluator cannot reliably determine whether a task is solvable or unsolvable, leading to variability in model performance due to this capability limitation.
Secondly, the stability of API status for a significant portion of online tools (55.6\% in ToolBench) is inconsistent. Users may be required to authorise the use of these tools or APIs, and tools provided by developers may be accessible during the initial construction of the benchmark but become unavailable later. This fluctuation further undermines the reliability and reproducibility of model performance assessments over time.
This situation results in a problem where the constructed queries in the benchmarks may no longer be completed with their originally referenced tools. Consequently, it is crucial to strike a balance between enhancing the stability of these benchmarks and maintaining their diversity and scope.

To address these issues, we propose a new benchmark named StableToolBench, which incorporates a virtual API system and a stable evaluation system. 
 We first build a virtual API system to replace the real one. As a start, we build a caching system to store the outputs of API calls. This approach ensures the stability and reproducibility of API behaviours. Given the limited number of benchmark questions, our caching system can cover a significant number of API call scenarios.
% Also, when the API behaviours of real APIs get changed, leading to calling failure, LLMs simulated APIs can serve as stable alternatives.
However, relying solely on a cache is insufficient because many APIs remain unavailable. To resolve this problem, we use large language models (LLMs) to simulate the behaviours of these APIs.
Specifically, we feed the documentation and few-shot real API calls if available in the cache to LLMs and ask LLMs to mock the behaviour of the APIs given a request. 
As a result, users can always get responses from APIs in an indistinguishable way as long as the LLMs are accessible. 
% Note that to mitigate the gap between the behaviours of real and simulated APIs, we use real API calls as the few-shot examples.
% With LLM simulation, behaviours of APIs can change when LLMs change their behaviour, especially when using a closed-source LLM such as GPT-4~\cite{openai2023gpt4}. We store calls to the API system 
% \color{black}
% Nevertheless, differences may exist between the behaviours of real and simulated APIs. To mitigate the gap, we create the simulator by feeding LLMs with API documentation and real API calls so that LLMs can mock real API behaviours well. 
% Experiments show that the simulated APIs can perform in an indistinguishable way from the real ones. 
% It is still crucial to highlight that, considering the irreplaceable nature of the reality provided by real APIs, our system prioritises real API calls.
% .\textcolor{red}{[Experiments follow]}
% Despite stability provided by LLMs, repeatedly calling LLMs can be very expensive. Responses generated by LLMs can also exhibit considerable variability. 
% Therefore, we additionally created an API response cache to store the results of real and simulated APIs. 
On the whole, our system first tries to find a hit in the cache.
Unless there is a cache miss and a real API call is not received, the simulated server will be used.
% The API call will not be directed to real and simulated APIs unless there is a cache miss.
% It is still crucial to highlight that, considering the irreplaceable nature of the reality provided by real APIs, our system prioritises real API calls. Only when a real API call is not received, the simulated server will be used.
% \textbf{(2) The stable evaluation system}. 

We then improve the evaluation system to make it more stable. We design two metrics (i.e., SoPR and SoWR) after judging solvable tasks and replace all the automatic evaluators with \texttt{GPT-4} to mitigate the randomness and indistinguishability during evaluation.
Experiments demonstrate that our virtual API system, when combined with the improved evaluation system, can provide stable evaluation against API modifications. Furthermore, our system exhibits significant reliability in terms of realism, diversity, and documentation following accuracy.
% Experiments demonstrate that our virtual API system consistently maintains stability in the face of real API modifications. Furthermore, our evaluation system yields more stable results and exhibits a high level of concordance with human assessments.
% Experiments show that our virtual API system can provide consistent stability across real API changes. In addition, our evaluation systems provides more stable results and achieves high agreement with human.
% [More Description] This makes it possible to curate real and diverse queries across different scenarios based on a large scale of APIs. [More Description] We also propose to a query-answer based NLI evaluation to better fit the real-world queries.

\begin{figure}[t!]
    \centering
    \includegraphics[width=\linewidth]{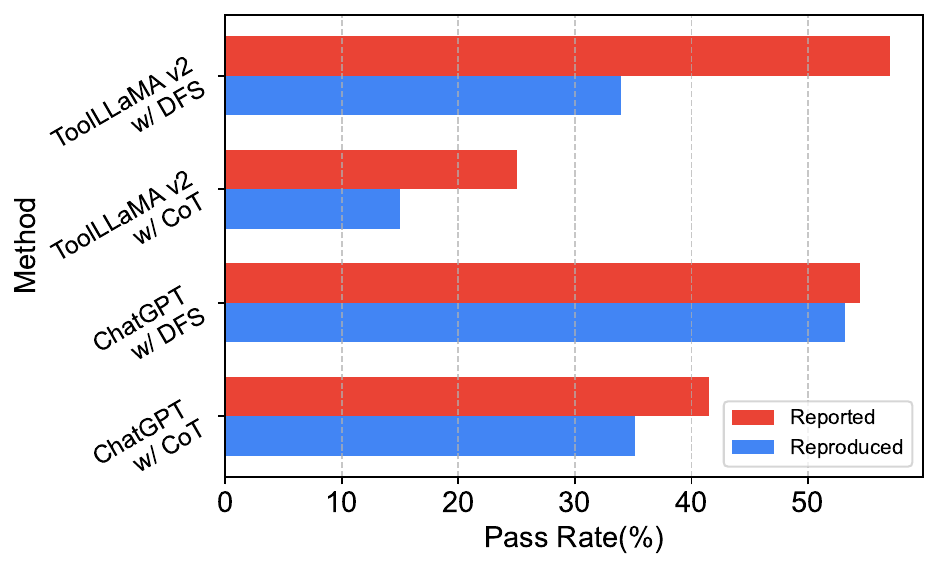}
    \caption{Comparison of performance (Pass Rate) reported in the paper and reproduced by us on the I1-Instruction group of ToolBench.}
    \label{fig:performance_comparison_failure}
\end{figure}
The main contributions of our work are summarised as follows:
\begin{itemize}[noitemsep]
    \item A tool-learning benchmark featured a large number of cached stable simulated APIs, well-balancing stability and reality of the APIs and much more stable evaluation metrics.
    \item Extensive experiments show that our benchmark provides much more stable model performance, robust to various types of API failures.
    \item Besides enhanced stability, our virtual API system exhibits reality, diversity and reliability comparable to that of the real API system.
\end{itemize}

\section{Stability Analysis on ToolBench}
In this section, we initiate a comprehensive analysis to reveal the stability of established tool benchmarks, using Toolbench~\citep{qin2023tool} as a case study.
We examine the stability of ToolBench by investigating three key dimensions: performance, evaluation, and API status.

\subsection{Stability of Performance}\label{sec:pre_analysis_performance}

Benchmarks are designed to consistently evaluate the performance of various models over time.
To test this consistency, we reproduce the model performances and record any variations.
Our study employs Chain-of-Thought (CoT;~\citealp{wei2023chainofthought}) and Depth First Search (DFS) strategies, leveraging ChatGPT and ToolLLaMA for comparative analysis.
We adhere strictly to the configurations detailed in ToolBench, utilising the ChatGPT version \texttt{gpt-3.5-turbo-0613} and \texttt{ToolLLaMA-v2}.
As depicted in \Cref{fig:performance_comparison_failure}, we compare the original Pass Rates for the I1-Instruction group reported by ToolBench with the Pass Rates we reproduced for four conventional methods.
\begin{figure}[]
    \centering
    \includegraphics[width=\linewidth]{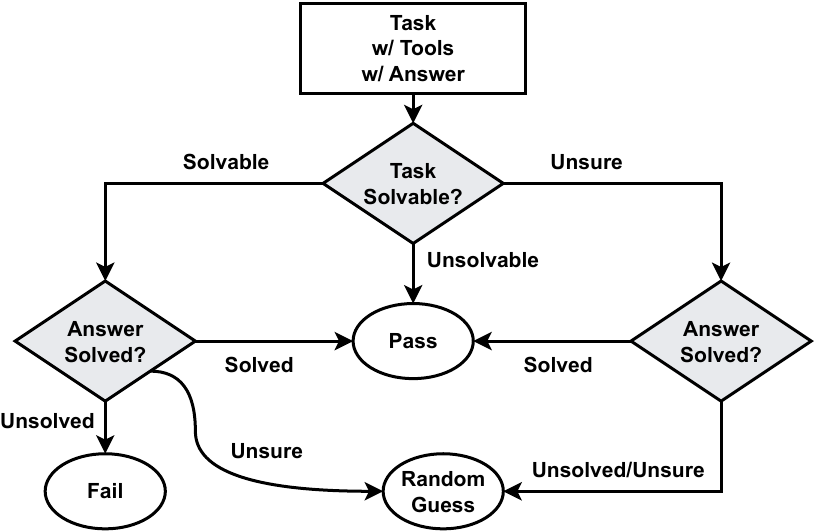}
    \caption{Pass Rate evaluation in ToolBench paper.}
    \label{fig:pass-rate}
\end{figure}
Our findings indicate a notable decline in the performance of all methods over time, which raises concerns about the stability of ToolBench as a benchmark.

% \begin{figure*}
%     % \centering
%     \includegraphics[width=\textwidth]{figs/table_2_perf_comparison.pdf} 
%     \caption{Comparison of performance (Pass Rate) reported in the paper and reproduced by us of ChatGPT and ToolLLaMA v2 on the I1-Instruction group of ToolBench.}
%     \label{tab:performance_comparison}
% \end{figure*}

\subsection{Stability of Evaluation}
\label{sta_eval}

In ToolBench, there are two types of metrics, including Pass Rate (PR) and Win Rate (WR). 
% PR is automatically evaluated by \texttt{gpt-3.5-turbo-16k}.
PR is calculated based on using \texttt{gpt-3.5-turbo-16k} to determine if a task is solvable and whether the generated answer can solve the corresponding task.
% PR is then calculated based on the solvability of the task and the effectiveness of the answer.
% The statuses of tasks are categorized as solvable, unsolvable, or unsure, while the statuses of answers can be solved, unsolved, or unsure.
\Cref{fig:pass-rate} details the computation process of PR.
Specifically, a solvable task results in a pass if the answer is solved, a failure if unsolved, and is randomly determined if unsure.
For tasks deemed unsure, a pass is assigned only if the answer is solved; otherwise, a random outcome is chosen.
If a task is unsolvable, the result defaults to a pass regardless of the answer status.
Moreover, WR is derived from the comparative PR of paired candidates.
% as shown in~\Cref{todo}.
A candidate's WR increases by one each time it passes while the other fails.
In all other situations, WR relies on \texttt{gpt-3.5-turbo-16k} to automatically evaluate the paired candidates.

% \usepackage{multirow}
% \usepackage{multirow}
% \begin{table}[]
% \small
% \centering
% \begin{tabular}{l|l|l|c|l}
% \toprule
%  & \multicolumn{3}{c}{\textbf{Task Status}} \\
%  & Solvable & \multicolumn{1}{l}{Unsolvable} & Unsure \\
%  \cmidrule
% Solved & Passed & \multirow{3}{*}{Passed} & Passed \\
% Unsolved & Failed &  & \multirow{2}{*}{Random} \\
% Unsure & Random &  &\\
% \bottomrule
% \end{tabular}
% \end{table}

\begin{table}[t!]
    \small
    \centering
    \resizebox{\linewidth}{!}{
    \begin{tabular}{l|ccc|ccc|c|c}
    \toprule
    \multirow{2}{*}{\textbf{Method}} & \multicolumn{3}{c|}{\textbf{Task}} & \multicolumn{3}{c|}{\textbf{Answer}} & \multirow{2}{*}{\textbf{Pass}} & \multirow{2}{*}{\textbf{Win}} \\
    & S & US & UE & S & US & UE & & \\
    \midrule
        \multirow{3}{*}{CoT} & 168 & 23 & 9 & 19 & 170 & 11 & 33.0 & 50.0 \\
         & 165 & 29 & 6 & 16 & 174 & 10 & 31.5 & 46.5 \\
         & 151 & 40 & 9 & 20 & 167 & 13 & 37.5 & 53.0 \\
        \midrule
        \multirow{3}{*}{DFS} & 116 & 68 & 16 & 17 & 167 & 16 & 50.5 & 54.0 \\
         & 122 & 59 & 19 & 20 & 162 & 18 & 46.5 & 48.0 \\
         & 132 & 54 & 14 & 22 & 157 & 21 & 55.0 & 56.0 \\
    \bottomrule
    \end{tabular}}
    \caption{Experiments use \texttt{GPT-3.5-Turbo-0613} with CoT and DFS. S, US, and UE indicate solvable (solved), unsolvable (unsolved), and unsure. Pass and Win denote pass rate and win rate, respectively. Win rates are evaluated against the first run of CoT. This experiment is run on 4 Feb 2024.}
    \label{tab:sta_eval}
\end{table}

To assess the stability of evaluation, we perform both CoT and DFS using \texttt{gpt-3.5-turbo-0613} on the I1-Instruction group dataset. 
These analyses are conducted using the provided tools and repeat over three iterations each.
The resulting PR and WR are presented in~\Cref{tab:sta_eval}, with detailed task and answer items.
% Although DFS generally exhibits higher PR than CoT, the distribution of tasks and answers is illogical.
Despite PRs of DFS being generally higher than CoT, the contribution of the task is larger than the answer.
However, it is worth noting that the tasks are the same in both CoT and DFS, where their results are expected to be consistent.
On the contrary, the discrimination of answers between CoT and DFS is weak, where a considerable proportion are unsolved.
Moreover, WR does not reflect the same trend as PR, where the second run of WR in DFS (48.0) is even lower than the first run of CoT.
Therefore, all the phenomena reflect that \texttt{gpt-3.5-turbo-16k} can not assume the role of the automatic evaluator in tool learning, which will be discussed in~\Cref{sec:evaluator}.

% \begin{figure}[h!]
%     \centering  
%      \begin{subfigure}[b]{\linewidth}
%         \includegraphics[width=\linewidth]{figs/task_status_count.pdf}
%         \caption{Task Status}
%       \end{subfigure}
%       \begin{subfigure}[b]{\linewidth}
%     \includegraphics[width=\linewidth]{figs/answer_status_count.pdf}
%     \caption{Answer Status}
%       \end{subfigure}
    
%     \caption{Task and Answer status}
%     \label{fig:sta_eval}
% \end{figure}

% \begin{figure}[ht!]
%     \centering
%     \includegraphics[width=\linewidth]{figs/status_eval.pdf}
%     \caption{Task and Answer status}
%     \label{fig:sta_eval}
% \end{figure}

\subsection{Stability of API Status}
\label{api_status}
\begin{figure}[]
    \centering   
    \includegraphics[width=\linewidth]{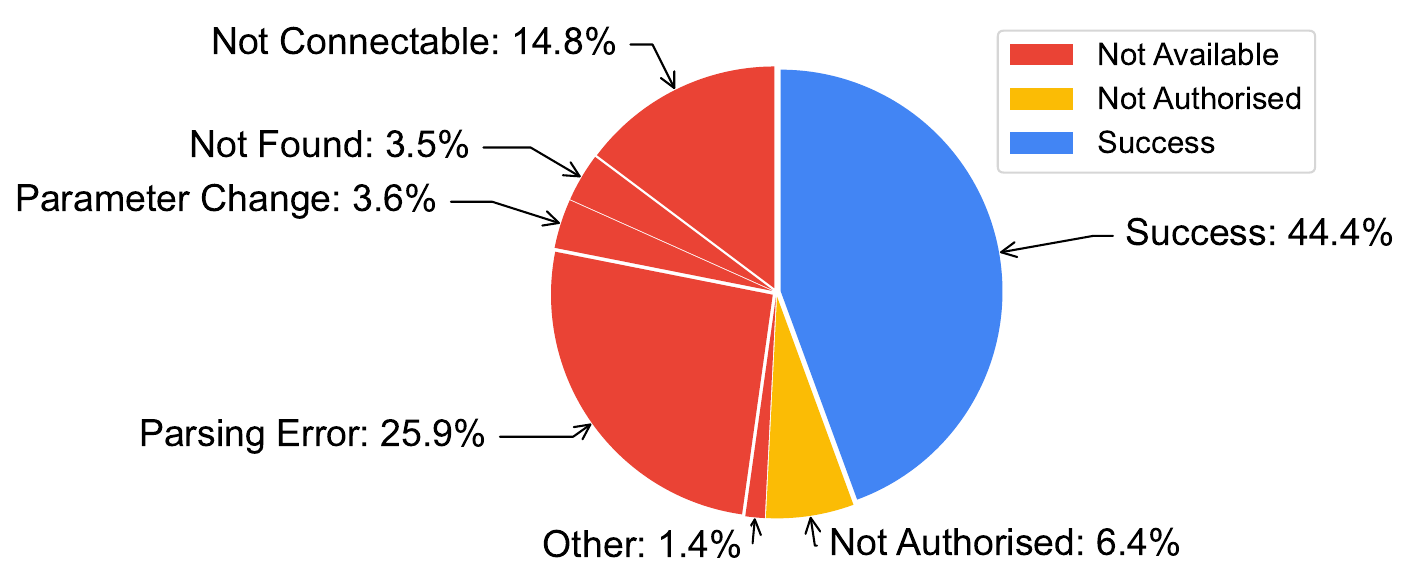}
    \caption{Statistics of API changes. Parsing errors are caused by post-processing errors of local API documentation, which have been solved in our benchmark.}
   
    \label{fig:api_change_info}
\end{figure}

% \textcolor{red}{Discuss the phenomenon that APIs are not callable in ToolBench and ToolEyes. These are resulting from authentication, documentation change, etc}.
% Results shown in \Cref{fig:toolbench-api-status}.
% To find the causes of the aforementioned instability,
We investigate the change of API status in ToolBench. 
In detail, we scan the original APIs downloaded from ToolBench, and use \texttt{gpt-4-turbo} to automatically write calls via the prompts as shown in \Cref{app:prompt_make_call}.
According to the keyword in API feedback, we classify these APIs into three categories: success, not availability, and not authorisation\footnote{Note that we use the \texttt{toolbench-key} provided in ToolBench to simulate the real running process in the benchmark.}.
The API status and the detailed errors of not availability are presented in \Cref{fig:api_change_info}. 
As can be seen, only 44.4\% of API calls are successful, while other API calls are mostly not available with various errors and some are not authorised.

% As for the not available APIs, xxx.

% authentication: 1058
% request_failure: 2426
% parameter: 591
% not_working: 248
% not_found: 583
% bugs: 4247

% \begin{table}[h!]
%     \centering
%     \small
%     \begin{tabular}{cc}
%      \toprule
%     \textbf{Status Type} & \textbf{Percentage} (\%) \\
%     \midrule
%     % No Change & 83.3 \\
%     % Parameter Change & 16.7 \\
%         No Change & 84.5 \\
%     Parameter Change & 15.5 \\
%      \bottomrule
%     \end{tabular}
%     \caption{APIs changed in ToolBench.}
%     \label{tab:api_change}
% \end{table}
\begin{figure}[h!]
    \centering
    \includegraphics[width=\linewidth]{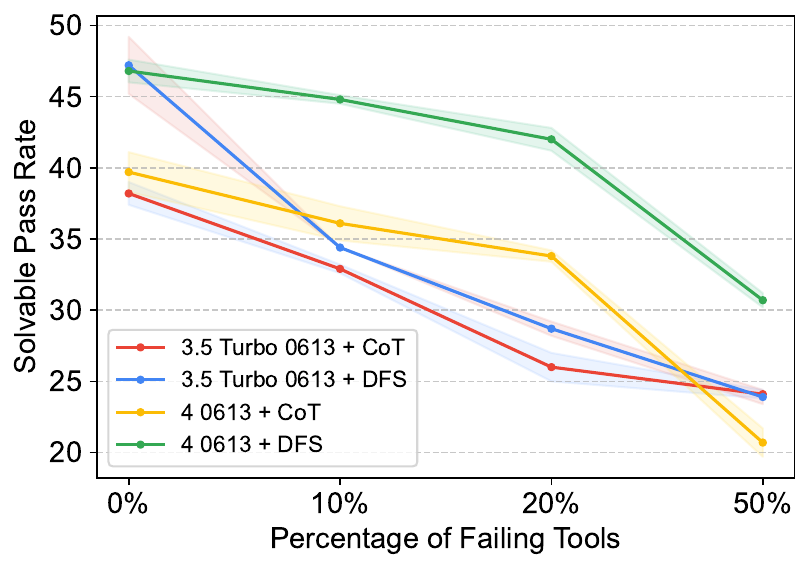}
    \caption{Solvable Pass Rate (SoPR) change when manually making APIs down on the I1 Instruction group.}
    \label{fig:real_api_stability_test}
\end{figure}

\begin{figure*}[t!]
    \centering
    \includegraphics[width=\textwidth]{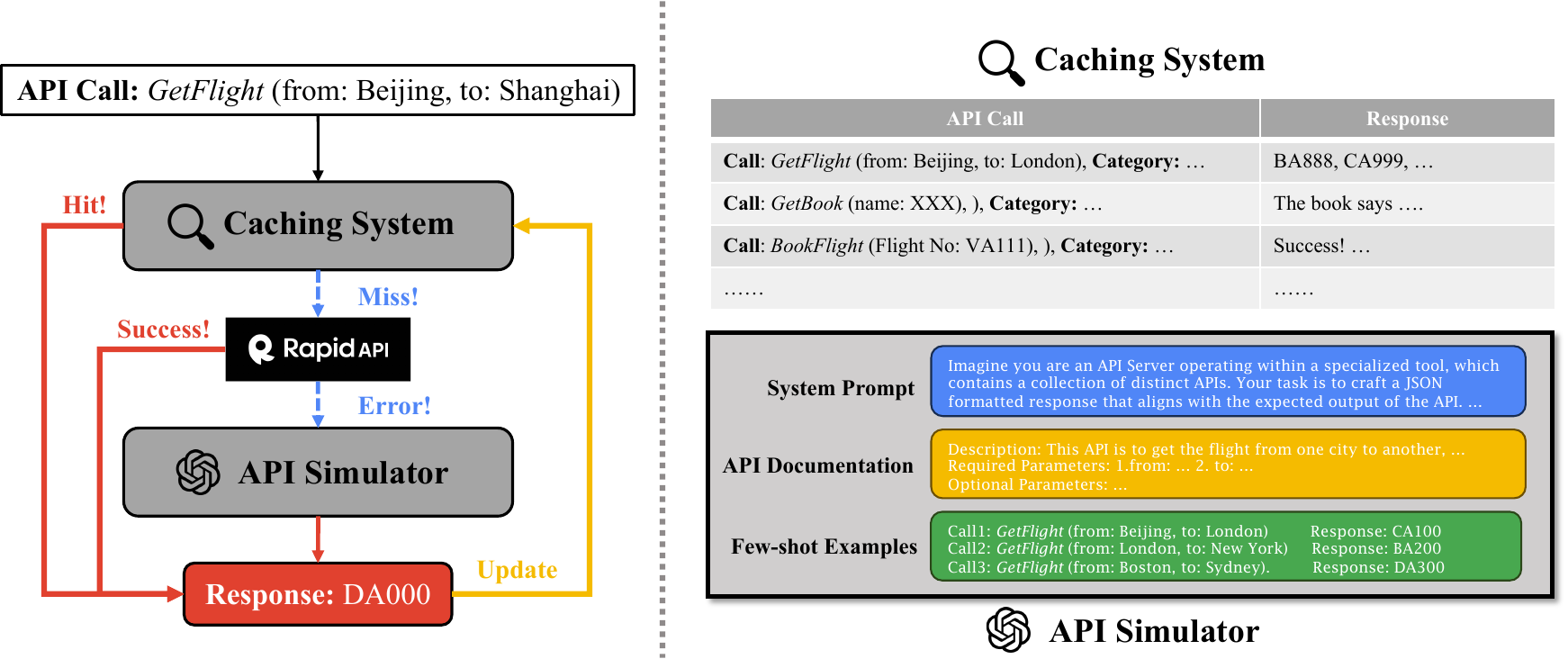} 
    \caption{The process of calling APIs in our proposed virtual API server.}
    \label{fig:main_figure}
\end{figure*}

% \begin{figure}[h!]
%     \centering
%     \includegraphics[width=\linewidth]{figs/real_solving_scores.pdf}
%     \caption{Solvable Pass Rate (SoPR) change when manually making APIs down on the I1 Instruction group.}
%     \label{fig:real_api_stability_test}
% \end{figure}
Furthermore, to validate the impact of API call failures on the stability of model performance, we manually make some success tools\footnote{In ToolBench, a tool is composed of several APIs. For example, a database tool can have two APIs: a writing API and a reading API. } down by returning a special failure call. Specifically, we randomly sample a proportion of success tools containing success APIs found in \Cref{fig:api_change_info}.
At testing time, when sampled tools are called, a special response will be thrown: \texttt{\{``error'': ``'', ``response'': ``This API did not return any useful information...''\}} to simulate the API call failures. 
We conduct baseline models with different proportions (i.e., 0\%, 10\%, 20\%, and 50\%) of sampled APIs on the I1 Instruction set.
% To confirm that the sampled tools will be called, we calculate the proportions of ground truth tools appearing in the sampled tools. 
% The results are shown in \Cref{app:proportion_sampling}.
Due to the issues in evaluation, we use our stable evaluation system proposed in~\Cref{sec:evaluation_system} as the same as our main experiments.
For each experiment, we evaluate three times and report the average scores as shown in \Cref{fig:real_api_stability_test}.
It can be seen that the performance degrades a lot when the proportion of successful APIs is down, thus the impact of API status on stability is considerable.

\section{StableToolBench}

Considering that stability is a crucial feature of benchmarking, in this paper, we specifically design a virtual API server and stable evaluation system to improve the stability based on ToolBench, and propose a new benchmark, named StableToolBench.

\subsection{Virtual API Server}

With real APIs, many of the failures encountered when reproducing its experiments are caused by expired APIs, network issues, or failed authentication.
To address this problem, we specifically propose a virtual API server with two components as illustrated in~\Cref{fig:main_figure}, including a caching system and API simulator.
Moreover, we design API calling rules to combine these two components to ensure the virtual API server is stable.

\paragraph{Caching System.}

We first implement a caching system that stores responses of all API callings to ensure consistency.
The caching system uses keys composed of their category, tool, API name, and arguments.
As a start, we populate the initial cache using the API call history from the training data and the reproduced test data released in ToolBench\footnote{\url{https://drive.google.com/drive/folders/1yBUQ732mPu-KclJnuQELEhtKakdXFc3J}}.
% We also update the cache with the reproduced test data released in ToolBench.
% New responses will be continuously recorded when running the experiment.
To ensure the quality of cached APIs, only valid records following the rule in \Cref{app:filter_rule} will be saved.
It is worth noting that we will also reserve some APIs with exceptions to keep the reality.
In this way, most API responses will be readily available, allowing the benchmark focus on probing the tool usage ability of designed methods with minimal impact on tool availability.
Furthermore, the API call in new experiments will also be continuously updated in the cache to ensure scalability.
The statistics of cache are shown in \Cref{tab:cache_components}.
\begin{table}[h!]
    \centering
    \small
    \begin{tabular}{lcccc}
        \toprule
         \textbf{Source} & \textbf{Train Set} & \textbf{Test Set} & \textbf{New Exp} & \textbf{Total} \\
         \midrule
         \textbf{Before} & 58,105 & 5,921 & 255,828 & 352,630 \\
         \textbf{After} & 25,995 & 2,393 & 136,592 & 164,980 \\
         \bottomrule
    \end{tabular}
    \caption{Cache components and their sizes before and after filtration. The cache of new experiments is updated until 12 Feb 2024.}
    \label{tab:cache_components}
\end{table}
Additionally, as an extra benefit, this approach reduces the latency introduced by interacting with real APIs, and also saves the costs for the API simulator discussed below.

\paragraph{API Simulator.}
Due to the limited coverage of the caching system, we propose to use LLMs to simulate API responses that are not in the cache and unavailable. 
Specifically, we ask \texttt{gpt-4-turbo} to simulate the API behaviour based on the original documentation in ToolBench.
The API documentation includes the descriptions of the functions and their corresponding parameters. 
% incorporate the documentation in the prompt of GPT-4 to generate a response given a requested input.
To mitigate the difference between simulated and real APIs, we use real API calls in the caching system as few-shot examples~\citep{brown2020language} for the LLM to better mock the behaviours. 
% These examples are sampled from previous calls on the same API stored in the caching system.
We keep the maximum number of examples at five.
When less than five examples exist in the cache, all of them will be used.
Detailed prompts can be found in \Cref{app:prompt_simulation}.
\begin{figure}[t!]
    \centering
    \includegraphics[width=\linewidth]{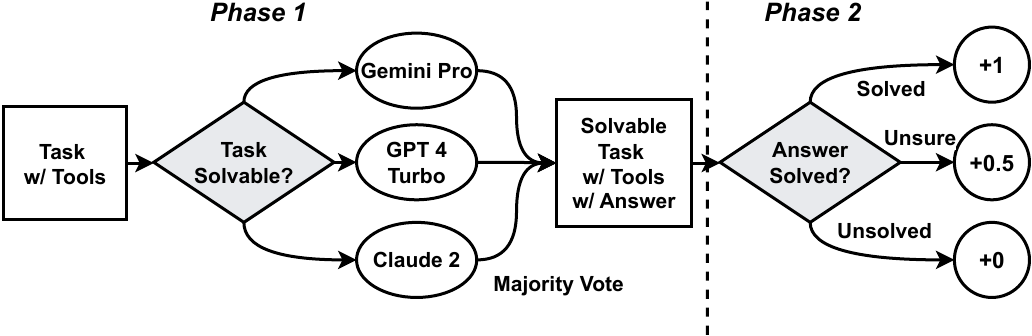}
    \caption{The process of our SoPR evaluation.}
    \label{fig:SoPR}
\end{figure}

\paragraph{API Calling Rules.}
Based on the caching system and the API simulator, we create API calling rules to ensure the stability of the virtual API server.
When a call request $(\mathrm{e},\mathrm{args})$, where $\mathrm{e}$ is the API endpoint and $\mathrm{args}$ is the arguments for that endpoint, is received, the system will first search the caching system for $(\mathrm{e},\mathrm{args})$ pair.
If a cache hit exists, the cached response will be directly returned.
When there is a cache miss, then the system will try to call the real API for a response to maintain the reality of the whole system.
If the real API calling is successful, the response will be returned.
However, when the caching system does not contain and the real API is not available, the system will finally call the simulated API. 
The final response whether from the real API or the simulated API will be saved to update in the caching system.
% Note that this will not affect the performance because all models evaluated will use the same response to a call whenever the the first call is made.

\subsection{Stable Evaluation System}
\label{sec:evaluation_system}

In this section, we propose a two-phase evaluation process, including judging solvable tasks and evaluating with two metrics, as shown in~\Cref{fig:SoPR}. Moreover, we replace all the automatic evaluators with \texttt{gpt-4-turbo}.

\paragraph{Judging Solvable Tasks.}\label{sec:judge_solvable_tasks}
\begin{table}[t!]
  \centering
  \small
  \resizebox{\linewidth}{!}{
  \begin{tabular}{lccccccc}
    \toprule
    & \textbf{I1-I} & \textbf{I1-C} & \textbf{I1-T} & \textbf{I2-I} & \textbf{I2-C} & \textbf{I3-I} & \textbf{Total}\\
    \midrule
    \textbf{Full} & 200 & 200 & 200 & 200 & 200 & 100 & 1,100 \\
    \textbf{Solvable} & 163 & 153 & 158 & 106 & 124 & 61 & 765\\
    % \midrule
    % \textbf{Total} & \multicolumn{2}{c}{765} & \multicolumn{3}{c}{1100} \\
    \bottomrule
  \end{tabular}}
  \caption{Statistics of original full and solvable tasks before and after judging. \texttt{C,I,T} stands for the Category, Instruction and Tool subgroup of the test set. Experiments below follow the denotation.}
  \label{tab:task_filteration}
\end{table}
Considering that both unsolvable and unsure tasks would introduce enormous randomness while the results of tasks fluctuate wildly, we try to obtain a fixed collection of solvable tasks to eliminate the problems.
To achieve this, we use three state-of-the-art LLMs, i.e., \texttt{gpt-4-turbo}, \texttt{gemini-pro}, and \texttt{claude-2}, to determine whether a task is solvable or unsolvable.
The prompt is shown in \Cref{app:prompt_task_solvability}.
The task will be judged as solvable when more than two models evaluate it as solvable.
The statistics of solvable tasks across all test datasets in ToolBench are shown in \Cref{tab:task_filteration}. 
% All reported scores are on the solvable tasks in the following experiment section.

\begin{table*}[t!]
    \centering
    \small
    \resizebox{\textwidth}{!}{
    \begin{tabular}{lccccccc}
        \toprule
        \textbf{Method} & \textbf{I1 Instruction} & \textbf{I1 Category} & \textbf{I1 Tool} & \textbf{I2 Category} & \textbf{I2 Instruction} & \textbf{I3 Instruction} & \textbf{Average} \\
        \midrule
     3.5 0613 (C)  & 52.2{\tiny $\pm{1.1}$} & 47.3{\tiny $\pm{0.6}$} & 53.6{\tiny $\pm{1.3}$} & 42.5{\tiny $\pm{2.1}$} & 35.8{\tiny $\pm{2.0}$} & 48.1{\tiny $\pm{0.8}$} & 46.6{\tiny $\pm{1.3}$} \\
    3.5 0613 (D)  & 60.3{\tiny $\pm{1.3}$} & 66.2{\tiny $\pm{1.2}$} & 67.1{\tiny $\pm{0.0}$} & 59.1{\tiny $\pm{0.4}$} & 51.3{\tiny $\pm{1.2}$} & 73.8{\tiny $\pm{2.3}$} & 63.0{\tiny $\pm{1.1}$} \\
    
    4 0613 (C) & 45.5{\tiny $\pm{0.4}$} & 57.4{\tiny $\pm{0.3}$} & 48.8{\tiny $\pm{0.7}$} & 43.0{\tiny $\pm{0.7}$} & 46.5{\tiny $\pm{0.9}$} & 48.1{\tiny $\pm{1.5}$} & 48.2{\tiny $\pm{0.8}$} \\
    4 0613 (D) & 57.3{\tiny $\pm{0.6}$} & 57.3{\tiny $\pm{0.3}$} & 60.9{\tiny $\pm{1.0}$} & 57.9{\tiny $\pm{1.0}$} & 51.3{\tiny $\pm{0.8}$} & 66.4{\tiny $\pm{2.4}$} & 58.5{\tiny $\pm{1.0}$} \\
    T-LLaMA (C) & 32.3{\tiny $\pm{1.0}$} & 40.3{\tiny $\pm{0.8}$} & 36.7{\tiny $\pm{0.5}$} & 34.7{\tiny $\pm{0.7}$} & 25.2{\tiny $\pm{0.4}$} & 33.9{\tiny $\pm{1.5}$} & 33.9{\tiny $\pm{0.8}$} \\
    T-LLaMA (D) & 44.5{\tiny $\pm{0.9}$} & 49.6{\tiny $\pm{1.3}$} & 48.9{\tiny $\pm{2.7}$} & 50.8{\tiny $\pm{1.1}$} & 31.9{\tiny $\pm{1.9}$} & 53.6{\tiny $\pm{2.0}$} & 46.6{\tiny $\pm{1.7}$} \\
    \midrule
   3.5 1106 (C)& 50.4{\tiny $\pm{0.5}$} & 45.1{\tiny $\pm{1.4}$} & 50.8{\tiny $\pm{0.3}$} & 48.7{\tiny $\pm{0.8}$} & 42.1{\tiny $\pm{0.4}$} & 55.7{\tiny $\pm{0.0}$} & 48.8{\tiny $\pm{0.6}$} \\
    3.5 1106 (D) & 62.8{\tiny $\pm{0.3}$} & 63.9{\tiny $\pm{1.2}$} & 65.6{\tiny $\pm{0.3}$} & 56.5{\tiny $\pm{0.7}$} & 56.9{\tiny $\pm{1.2}$} & 67.2{\tiny $\pm{1.3}$} & 62.2{\tiny $\pm{0.8}$} \\
    4 Turbo (C) & 52.8{\tiny $\pm{1.3}$} & 56.6{\tiny $\pm{0.9}$} & 51.9{\tiny $\pm{0.5}$} & 51.9{\tiny $\pm{1.0}$} & 52.8{\tiny $\pm{0.8}$} & 52.5{\tiny $\pm{0.0}$} & 53.1{\tiny $\pm{0.8}$} \\
    4 Turbo (D) & 59.2{\tiny $\pm{0.5}$} & 61.7{\tiny $\pm{0.7}$} & 65.7{\tiny $\pm{1.0}$} & 55.6{\tiny $\pm{0.6}$} & 55.2{\tiny $\pm{0.4}$} & 66.1{\tiny $\pm{4.3}$} & 60.6{\tiny $\pm{1.3}$} \\

         \bottomrule
    \end{tabular}
    }
    \caption{Solvable pass rate scores. We run all models once, evaluate three times and take the average results. ``3.5 0613'', ``4 0613'', ``3.5 1106'', ``4 Turbo'', ``T-LLaMA'' stands for \texttt{gpt-3.5-turbo-0613}, \texttt{gpt-4-0613}, \texttt{gpt-3.5-turbo-1106}, \texttt{gpt-4-turbo-preview}, 
    \texttt{ToolLLaMA v2} respectively. C and D stand for CoT and DFS respectively. The experiments below follow the denotation. We use \texttt{gpt-4-turbo-2024-04-09} as the evaluator. Evaluation done on May 2024.}
    \label{tab:main_sopr}
\end{table*}

% \begin{table}[h!]
%     \centering
%     \small
%     \begin{tabular}{lcc}
%         \toprule
%         \multirow{2}{*}{\textbf{Group}} & \textbf{After} & \textbf{Before} \\
%         & \textbf{Filtration} & \textbf{Filtration}\\
%         \midrule
%         {I1 Instruction} & 163 & 200\\
%         {I1 Category }& 153 & 200 \\
%         {I1 Tool} & 158 & 200 \\
%         {I2 Instruction} & 106 & 200\\
%         {I2 Category} & 124 & 200 \\
%         {I3 Instruction} & 61 & 100 \\
%         \midrule
%         {{Total}} & 765& 1100\\
%         \bottomrule
%     \end{tabular}
%     \caption{Query count statistics before and after filtration. \csj{todo}}
%     \label{tab:task_filteration}
% \end{table}

\paragraph{Metrics.}
We then report Solvable Pass Rate (SoPR) and Solvable Win Rate (SoWR) based on our obtained solvable tasks.
Due to the limitation of \texttt{gpt-3.5-turbo-16k} in tool learning, we uniformly adopt \texttt{gpt-4-turbo} as the automatic evaluator.
SoPR is in essence PR with all tasks solvable and only assesses the answers using the same prompt in ToolBench.
The evaluator assigns outcomes of answers categorised as Solved, Unsolved, or Unsure, which respectively contribute scores of 1, 0.5, and 0 to the overall SoPR calculation.
As for SoWR, when one is solved and the other is unsolved, the solved one wins.
Under other circumstances, \texttt{gpt-4-turbo} will be used to make a win-lose decision.  
% Note that the evaluation will be made whenever an \texttt{Unsure} label occurs.
% When calculating SoPR, the task description, available tools are fed into the automatic evaluator, using the same prompt as that in evaluating answer status when calculating Pass Rate\footnote{Prompts used to evaluate Pass Rate and Win Rate from ToolBench are at \url{https://github.com/OpenBMB/ToolBench/blob/master/toolbench/tooleval/evaluators/tooleval_gpt-3.5-turbo_default/template.txt}}.
% Furthermore, we improved the Win Rate with our SoPR to make it more stable, named Solvable Win Rate (SoWR). 
% Originally, when two methods received a \texttt{Passed} label and a \texttt{Failed} label during Pass Rate evaluations, the one with \texttt{Passed} label always wins when making Win Rate evaluations. Under other circumstances, an LLM based automatic evaluator will be used to make a win-lose decision. 

% \begin{table}[]
%     \centering
%     \small
%     \begin{tabular}{c|cc}
%         \toprule
%         \textbf{Methods} &  \\
%         \midrule
%          GPT 4 Turbo & 74.0 & 78.0\\
%          GPT 3.5 Turbo & 68.0 & 56.0 \\
%          \bottomrule
%     \end{tabular}
%     \caption{Human evaluation on answer solving (for pass rate) and comparison (for win rate).}
%     \label{tab:human_eval_task}
% \end{table}

\section{Experiment}

\subsection{Performance}\label{sec:performance}

% \textcolor{red}{Run different baseline models on the benchmark. The goal is to show that ranking of models on the benchmarks is the same as the precedents, i.e. improved stability will not change discrimination of the benchmark.}
Following ToolBench, we run \texttt{gpt-3.5-0613}, \texttt{gpt-4-0613}, \texttt{ToolLLaMA-v2} with CoT and DFS, replenishing with latest models \texttt{gpt-3.5-1106} and \texttt{gpt-4-turbo}.  
The results of SoPR and SoWR are shown in \Cref{tab:main_sopr,tab:main_sowr}. Generally, GPT-4 series models outperform GPT-3.5 models, while ToolLLaMA performs worst with the same inference algorithm.
Also, DFS significantly outperforms CoT whichever LLMs are used.
These phenomena are consistent with ToolBench.
Furthermore, probably thanks to the improved function calling capabilities, newer GPT models performed better. 
% Nevertheless, with more stable tools, gaps between LLMs are smaller with DFS than with CoT. This may indicate that more trials can significantly boost problem-solving performance.
\begin{table}[t!]
    \centering
    \small
    \resizebox{\linewidth}{!}{
    \begin{tabular}{lccccccc}
        \toprule
        \textbf{Method} & \textbf{I1-I} & \textbf{I1-C} & \textbf{I1-T} & \textbf{I2-I} & \textbf{I2-C} & \textbf{I3-I} & \textbf{Avg} \\
        \midrule
        3.5 0613 (D) & 60.7 & 67.3 & 59.5 & 63.2 & 62.1 & 75.4 & 64.7 \\
        4 0613 (C) & 54.6 & 58.8 & 58.2 &  75.5 & 60.5 & 62.3 & 61.7 \\
        4 0613 (D) & 62.6 & 62.7 & 58.2 & 74.5 &62.9 &  67.2 & 64.7 \\
        T-LLaMA (C) & 31.3 &28.1 &  33.5 & 35.8 & 33.9 &  24.6 & 31.2 \\
        T-LLaMA (D) & 44.8 & 45.8 & 44.3 & 59.4 & 41.1 & 50.8 & 47.7 \\
        \midrule
        3.5 1106 (C) &  47.2 & 47.7 & 44.9 & 50.9 & 54.0 & 62.3 & 51.2 \\
        3.5 1106 (D) & 55.8 &53.6 &  51.9 & 68.9 & 59.7 & 68.9 & 59.8 \\
        4 Turbo (C) & 71.2 &{77.1} &  61.4 & { 79.2} & {71.8} & {67.2} & {71.3} \\
        4 Turbo (D) & {73.0} &{75.2} &  {68.4} &  {77.4} & {66.9} & {60.7} & {70.2} \\
        \bottomrule
    \end{tabular}}
    \caption{Solvable Win Rate scores. We run all models once against \texttt{GPT-3.5-Turbo-0613 + CoT} and evaluate three times. We follow the ToolBench implementation to take the most frequent result for each query during evaluation. The experiments below follow the denotation. We use \texttt{gpt-4-turbo-2024-04-09} as the evaluator. Evaluation done on May 2024.}
    \label{tab:main_sowr}
\end{table}

\subsection{Stability of Virtual API Server}
% \textcolor{red}{Manually make ... of the APIs down (set probability of each calling APIs ), report the performance change of different methods. Aim to show the instability of the original system and stability of our system.} Results are shown in 
\begin{figure}[h!]
    \centering
    \includegraphics[width=\linewidth]{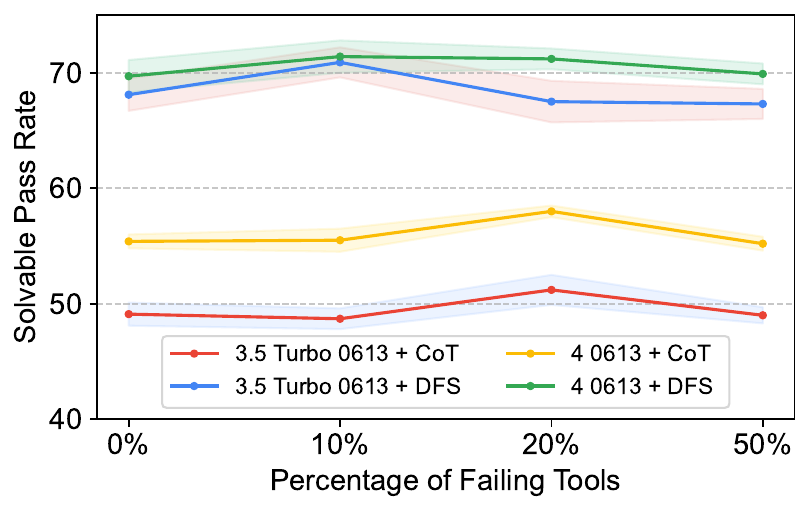}
    \caption{Performance change when manually making APIs down with our virtual online API system. The results are averaged over all six groups. Solving rates are reported. We run each experiment one time and evaluate it three times and take the average score. Unless otherwise stated, \texttt{gpt-4-turbo-preview} at the time of testing is used in this section. This experiment was done in Feb 2024.}
    \label{fig:simulated_api_stability_test}
\end{figure}

Following the same setups as in \Cref{api_status}, we manually make the same success tools not available during the running time.
In our design, when a call is on an unavailable tool, it will be directed to the simulated API immediately.
Compared to \Cref{api_status}, the results as shown in \Cref{fig:simulated_api_stability_test} are much more stable with our virtual API server.
Even when 50\% of APIs are not available, changes in performance are still not significant, which is explainable within the range of variance.

% \textcolor{red}{change gpt-4 versions and other hyperparameters and run multiple sets of experiments. Aim to show change of GPT-4 won't affect the stability significantly.} 
Considering we use \texttt{gpt-4-turbo} as the backbone of the API simulator which may change even with the same version number, we ablate different versions and different temperatures of \texttt{gpt-4-turbo}.
The results are shown in \Cref{tab:server_config}.
Under different settings of the backbone LLMs, the performance change is still acceptable within the variance of LLM evaluation, indicating the robustness of our API simulators.

% \textcolor{red}{Discussion: An ultimate solution is to train a open-source models that make mock the API behavior well.
% }

% \begin{figure}
%     \centering
%     \includegraphics[width=\linewidth]{figs/virtual_solving_scores_against_server.pdf}
%     \caption{Performance of baselines with different settings of the LLM server. The results are averaged over all six groups. Solving rates are reported. We run each experiment one time and evaluate three times and take the average score.}
%     \label{fig:server_config}
% \end{figure}

\begin{table}
    \centering
    \small
    \resizebox{\linewidth}{!}{

    \begin{tabular}{lcccc}
        \toprule
        \multirow{2}{*}{\textbf{GPT-4 Config}} & \multicolumn{2}{c}{\textbf{1106-preview}} &  \multicolumn{2}{c}{\textbf{0125}} \\
        & \texttt{T=1} & \texttt{T=0.1} & \texttt{T=1} & \texttt{T=0.1}\\
        \midrule
         3.5 0613 (C) & 49.1{\tiny $\pm{1.0}$}  & 49.0{\tiny $\pm{0.8}$} & 52.1{\tiny $\pm{0.5}$} & 50.2{\tiny $\pm{0.8}$} \\
         3.5 0613 (D) & 68.1{\tiny $\pm{1.4}$}  & 67.4{\tiny $\pm{0.9}$} & 69.3{\tiny $\pm{1.0}$} & 67.9{\tiny $\pm{1.2}$}\\
         4 0613 (C) & 55.4{\tiny $\pm{0.6}$}  & 56.3{\tiny $\pm{0.7}$} & 57.7{\tiny $\pm{0.5}$} & 54.5{\tiny $\pm{0.6}$} \\
         4 0613 (D) & 69.7{\tiny $\pm{1.4}$}  & 70.3{\tiny $\pm{1.0}$} & 71.4{\tiny $\pm{0.7}$} & 70.4{\tiny $\pm{1.3}$}\\
         \midrule
         3.5 1106 (C) & 52.1{\tiny $\pm{0.7}$}  & 51.1{\tiny $\pm{0.5}$} & 54.5{\tiny $\pm{0.9}$} & 52.7{\tiny $\pm{0.6}$} \\
         3.5 1106 (D)  & 69.9{\tiny $\pm{0.7}$}  & 71.2{\tiny $\pm{0.9}$} & 70.0{\tiny $\pm{0.9}$} & 71.0{\tiny $\pm{0.9}$}\\
         4 Turbo (C) & 60.8{\tiny $\pm{0.7}$}  & 62.4{\tiny $\pm{0.8}$} & 63.6{\tiny $\pm{0.4}$} & 64.0{\tiny $\pm{1.1}$} \\
         4 Turbo (D) & 73.2{\tiny $\pm{1.1}$}  & 76.2{\tiny $\pm{0.9}$} & 75.0{\tiny $\pm{0.7}$} & 77.3{\tiny $\pm{0.9}$}\\     
        \bottomrule
    \end{tabular}
    }
    \caption{Performance of baselines with different settings of the LLM server. Results are averaged over all groups and reported in SoPR. We run each experiment one time and evaluate three times and take the average score.}
    \label{tab:server_config}
\end{table}

% Different models, same input, same outputs, temp=0

\subsection{Turing Test of API Simulator}
% \textcolor{red}{Aiming at testing the reality of APIs. 
% Sampling 100 API calls from real API responses and LLM faked responses, and ask GPT-4 / Human whether the two are indistinguishable (guess which is model and which is real). If the results are around 50\%. Then the test is passed.}
To test the effectiveness of API simulators, we design a ``Turing Test''~\cite{Turing2009} between the real APIs and the simulated ones.
% To successfully run experiments on a tool-learning benchmark, the fundamental requirement for the API system is that the responses of the API system are usable and believable by the testing model. 
% In this way, the testing model can use the outputs to finish an instruction. 
Note that we believe it is not required for API simulators to exactly output the same answers as those of real APIs, where rationality is more important.
% This is because the exact reality of API responses is not required for task completion.
For example, when a query asks about the weather today, the API simulator does not need to retrieve the ``real'' temperature. Instead, the API simulator needs to generate a reasonable temperature number.

\begin{figure}[t!]
    \centering
    \includegraphics[width=\linewidth]{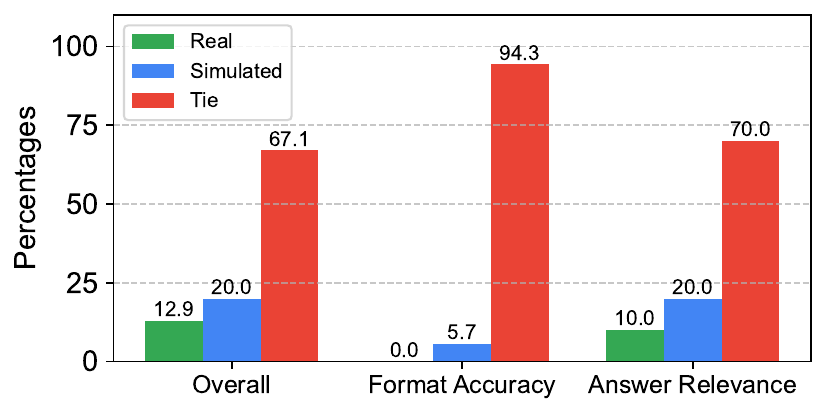}
    \caption{Results of the ``Turing Test'' for the real and simulated APIs. Results are win-lose-tie percentages.}
    \label{fig:turing_test}
\end{figure}

To do the test, we first sample 70 available real APIs and their corresponding simulated APIs. 
% We then manually filter out unsuccessful real API calls, resulting in 70 API calls.
Given the API callings and their real and simulated response pairs, we ask three human annotators to determine which response more closely resembles an actual API response overall, based on the given descriptions of the API functions.
We ask human annotators to evaluate along three dimensions: Overall, Format Accuracy and Answer Relevance.
The annotator first need to answer which response is overall more like a real response.
When assessing format accuracy, annotators must determine which response more accurately adheres to the format specifications outlined in the documentation. In evaluating answer relevance, they are tasked with identifying which response more effectively fulfills the instruction in accordance with the documentation's guidelines.
The results are shown in ~\Cref{fig:turing_test}.
Surprisingly, human annotators cannot distinguish simulated and real APIs very well, where the simulated APIs are judged to act more like real situations.
Moreover, the proportion of tie is much larger, indicating that simulated APIs can work very similarly to real APIs.
% To further analyze the reason, we further ask human annotators to distinguish two kinds of APIs in terms of format accuracy and answer relevance only. As shown in \Cref{fig:turing_test}, one can see simulated APIs can work very similarly to the real APIs.

% \begin{table*}
%     \centering
%     \small
%     \begin{tabular}{cccccccccc}
%         \toprule
%         & \multicolumn{3}{c}{\textbf{Overall}} & \multicolumn{3}{c}{\textbf{Format Accuracy}}& \multicolumn{3}{c}{\textbf{Answer Relevance}}\\
%         \cmidrule(lr){2-4}
%         \cmidrule(lr){5-7}
%         \cmidrule(lr){8-10}
%         & {Real} &  {Simulated} & {Tie} & {Real} &  {Simulated} & {Tie}& {Real} &  {Simulated} & {Tie} \\
%         \midrule
%         % GPT-4-Turbo & 31.0 & 62.1 & 6.9 & 34.5 & 58.6 & 6.9 & 37.9 & 55.2 & 6.9  \\
%         % Gemini Pro & 31.0 & 69.0 & 0.0 & 20.7 & 79.3 & 0.0 & 31.0 & 65.5 & 3.4 \\
%         Human & 12.9\% & 20\% & 67.1\% & 0 \% & 5.7\% & 94.3\% & 10\% & 20\% & 70\% \\
%         \bottomrule
%     \end{tabular}
%     \caption{Results of the ``Turing Test'' for the real and LLM simulated APIs. \textcolor{red}{TODO: Change to bar figure.}}
%     \label{tab:turing_test}
% \end{table*}

% We sample 50 LLM simulations from the Turing Test questions discussed in Section 4.3. Then we invite a human evaluator to answer whether the LLM simulation follows the documentation, i.e., whether it is a reasonable simulation that obeys the setting of the documentation. Results show that 90% of the simulations are deemed to be reasonable, 6% are not reasonable and 4% are not sure. It shows that the simulated APIs can well follow documentation.

In addition to the Turing Test mentioned above, we assess the quality of LLM simulations by evaluating the adherence of simulated outputs to their corresponding documentation. To conduct this evaluation, we randomly sample 50 simulated outputs along with their documentation from the Turing Test dataset. A human evaluator is then tasked with determining whether the LLM simulations reasonably follow the provided documentation. The results indicate that 90\% of the simulations are deemed reasonable, 6\% are considered unreasonable, and 4\% are uncertain. These findings suggest that the LLM is highly capable of generating simulated responses that adhere closely to the provided documentation.

% \begin{itemize}
%     \item Different inputs -> different outputs, same formats
% \end{itemize}

\subsection{Diversity of API Simulator}
\begin{figure}[h!]
    \centering
    \includegraphics[width=\linewidth]{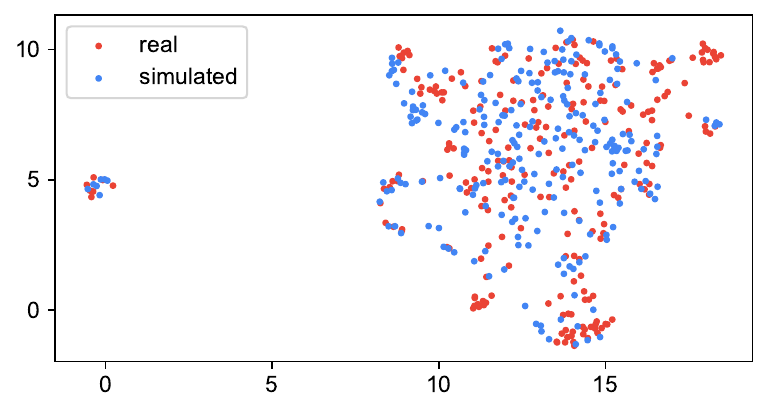}
    \caption{Visualisation of the embeddings of responses from real and simulated APIs.}
    \label{fig:diversity_comparison}
\end{figure}

% \textcolor{red}{Aim to show that API calls are diverse enough, so that the effective numbers of APIs won't decrease with simulated APIs.}
% \begin{enumerate}
%     \item Select a category of APIs, call these APIs in real and simulated ways. Calculate embeddings of these calls and see clustering outputs.
%     \item Call a selected APIs (potentially 10) with various arguments. See the difference across different arguments for both real and simulated APIs.
% \end{enumerate}
With the LLM simulation, API simulators will not exactly feedback the same as real APIs. 
Hence, a natural concern is whether the simulated APIs will degrade in diversity in API functionalities.
To study the problem, firstly, we explore the distribution of real and simulated API responses.
We first use all 246 APIs in the \texttt{Tool} API category from the successful APIs mentioned in \Cref{api_status}.
Then, we use the same call arguments to call these real and simulated APIs.
All the responses are encoded using S-BERT~\cite{reimers-2019-sentence-bert} and their corresponding embeddings are visualised by UMAP~\cite{mcinnes2018umap-software}. 
Detailed configuration is shown in \Cref{app:diversity_conf}.
The result is shown in \Cref{fig:diversity_comparison}. As can be seen from the figure, real and simulated APIs occupied similar embedding space, indicating that the diversity of simulated APIs is similar to the real APIs.

Secondly, we try to explore the behaviour when a simulated API is given several calls with different input arguments. 
We sample 60 APIs from all successful APIs and make 5 different calls to each API, using the same prompt as in \Cref{app:prompt_make_call}.
We then count the number of APIs that give exactly the same responses in any 2 of the 5 calls. Results show that only 2 of 60 APIs contain such responses, which supports the sufficient diversity of our simulated APIs.

\subsection{Effectiveness of Caching System}
% We run experiments on different methods and record the cache hit rate.
To show the effectiveness of our caching system in maintaining the stability of the virtual API server, we run several methods and record the cache hit rates. In detail, we run four methods used in ToolBench, \texttt{gpt-3.5-turbo-0613} and \texttt{gpt-4-0613} with CoT and DFS. Reproduction data in ToolBench of these methods has been used in the cache. Results are recorded in~\Cref{tab:cache_hit}, which shows that rerunning these in-domain methods has a very high cache hit rate. This means that most of the call responses are fixed and instability from the API system are much smaller. Nevertheless, models and methods may change over time, and therefore, we further run \texttt{gpt-3.5-turbo-1106} and \texttt{gpt-4-turbo-preview} with CoT and DFS.  As can be seen in the table, although the cache hit rates are smaller with these out-of-domain methods, the scores are still high enough to mitigate the instability significantly.

\subsection{Human Evaluation of Evaluator}
\label{sec:evaluator}

Considering that GPT-3.5 is limited to evaluating the performance in tool learning, we replace the automatic evaluators with stronger LLMs.
In this section, we manually assess the correctness of different automatic evaluators. 
Here, we sample 100 task-solvable questions, 50 answer-solving questions in the PR / SoPR evaluation, and 50 comparison questions in the WR / SoWR evaluation from the experiments running.

\begin{table}
    \centering
    \small
    \begin{tabular}{lccc}
        \toprule
        \textbf{Methods} & \textbf{Final} & \textbf{Mid} & \textbf{Start}\\
        \midrule
        GPT 3.5 Turbo 0613 + CoT & 96.7 & 36.2 & 11.7 \\
        GPT 3.5 Turbo 0613 + DFS & 97.0 & 34.5 & 11.6 \\
        GPT 4 0613 + CoT &  96.5 & 36.2 & 11.7\\
        GPT 4 0613  + DFS & 97.0 & 35.0 & 11.7 \\
         \midrule 
        GPT 3.5 Turbo 1106 + CoT & 91.4 & 35.1 & 11.8\\
         GPT 3.5 Turbo 1106 + DFS & 75.8 & 34.5 & 11.4 \\
        GPT 4 Turbo + CoT & 88.2 & 35.0 & 11.6 \\
        GPT 4 Turbo + DFS & 77.8 & 34.5 & 11.8\\
        \bottomrule
    \end{tabular}
    \caption{Cache hit rate (\%) with various models and methods. Final, Mid, and Start represent the final version of the cache, the mid-way version containing 151,152 (91.6\%) items of the final version, and the starting version containing only the train and test set. Experiments are independent runs of \Cref{sec:performance} with fixed cache, run on 13 Feb 2024.}
    \label{tab:cache_hit}
\end{table}
We then collect all the corresponding answers of different LLMs during previous evaluations.
% We use the same prompt and format as in the experiments. 
% We run \texttt{gpt-3.5-turbo}, \texttt{gpt-4-turbo}, \texttt{gemini-pro}, and \texttt{claude-2} on task solvable questions and \texttt{gpt-3.5-turbo}, \texttt{gpt-4-turbo} on answer solving and comparison questions.
These questions are further manually labeled by three human annotators to obtain the ground truth.
With the ground truth, we calculate the accuracy scores of these models and show the results in \Cref{tab:human_eval_task}. 
It can be seen that our used LLMs (i.e., Claude 2, Gemini, and GPT-4) are much better than GPT-3.5 in ToolBench to determine the solvability of tasks, where the Gemini and GPT-4 outperform by a large margin.
In both the evaluation of answers and comparison, GPT-4 significantly outperforms GPT-3.5, especially in the comparison to compute WR.
It is worth noting that all the accuracies of GPT-3.5 are lower than 70\%, indicating that GPT-3.5 cannot assess the performance in tool learning.
% \texttt{gemini-pro} and \texttt{claude-2} perform much better than \texttt{gpt-3.5-turbo} as well, which supports our model choices when judging task solvability above. 
% As a result, we replace {gpt-3.5-turbo} with \texttt{gpt-4-turbo} in the experiments below.

% \begin{table*}[]
%     \centering
%     \small
%     \begin{tabular}{cccccccc}
%         \toprule
%         \textbf{Method} &G1 Cat & G1 Inst & G1 Tool & G2 Cat & G2 Inst & G3 Inst & Avg \\
%         \midrule
%          ChatGPT + CoT & \\
%          ChatGPT + DFS \\
%          GPT-4 + CoT \\
%          GPT-4 + DFS \\
%          ToolLLaMA + CoT \\
%          ToolLLaMA + DFS \\
%          \bottomrule
%     \end{tabular}
%     \caption{NLI score}
%     \label{tab:my_label}
% \end{table*}

% \subsection{Human Evaluation on the NLI Metric}

% \section{Methods}
% \input{sections/3_methods}

% \section{Experiments}
% \input{sections/4_results}

% \section{Experiments}
% \input{sections/4_exps}

% \section{Analysis}
% \input{sections/5_ablation}

\section{Related Work}

\paragraph{Tool Learning Benchmarks.}
% \csj{please rewrite.}
Recent studies have shed light on the burgeoning capabilities of LLMs in understanding and mastering tools~\citep{li2023apibank, patil2023gorilla, gpt4tools, song2023restgpt, tang2023toolalpaca, ye2024tooleyes, xu2023tool}.
Gaining access to external tools endows LLMs with real-time factual knowledge~\citep{yang2023chatgpt}, multimodal functionalities~\citep{gupta2023visual}, and specialised skills in vertical domains~\citep{jin2023genegpt}. 
However, few work has been done to explore the stability of the tool environment in specific benchmarks and how it affects the LLMs' performance in tool-augmented tasks. 
% However, open-source LLMs still lag far behind the state-of-the-art LLMs in tool use, and how tool-use ability is acquired by SOTA LLMs remains unclear. In this paper, we aim to push the boundary by creating a stable tool-learning benchmark.

\paragraph{Tool Inference Methods.}
Recent literature has begun to explore various methodologies for integrating tool functionalities within LLMs. 
Notably, the robust in-context learning prowess of LLMs, as demonstrated in \cite{brown2020language}, has facilitated the augmentation of LLMs with external tools via in-context tool descriptions and demonstrations ~\citep{hsieh2023tool, ruan2023tptu, mialon2023augmented}.
An alternative approach involves the explicit training of LLMs ~\citep{patil2023gorilla, tang2023toolalpaca, chen2023fireact, qin2023toolllm, huang2023metatool} using datasets enriched with tool interactions, thereby familiarising models with the nuances of tool usage. 
% This area of research is pivotal, as it directly influences the efficiency and effectiveness of tool-augmented LLMs in performing a wide array of tasks.

\begin{table}[t!]
    \centering
    \small
    \begin{tabular}{lccc}
        \toprule
        \textbf{Methods} & \textbf{Solvability} &\textbf{Solving} & \textbf{Comparison} \\
        \midrule
         Claude 2 & 71.0 & - & -\\
         Gemini Pro & \textbf{82.0} & - & -\\
         GPT 3.5 Turbo & 65.0 & 68.0 & 56.0 \\
         GPT 4 Turbo & 80.0 & \textbf{74.0} & \textbf{78.0}\\

         \bottomrule
    \end{tabular}
    \caption{Human evaluation on task solvability, answer solving (for pass rate) and comparison (for win rate).}
    \label{tab:human_eval_task}
\end{table}

\paragraph{Evaluation in Tool Learning.}
Evaluating the performance of LLMs in tool-augmented tasks presents unique challenges and opportunities.
Numerous works have been developed for the assessment of tool utilisation, primarily emphasising response comparison ~\citep{zhuang2023toolqa}, tool call accuracy ~\citep{patil2023gorilla}, or a synthesis of these aspects ~\citep{li2023apibank}. Distinguishing itself, \cite{qin2023toolllm} introduces an innovative methodology by integrating a large language model (LLM) as a judge to evaluate the comprehensive solution path. 
Subsequent research ~\citep{wang2023mint} focuses on the multi-turn interaction capabilities of LLMs with both tools and user feedback.
In a departure from the aforementioned approaches, \cite{chen2023teval} presents itself as the inaugural benchmark specifically tailored for the fine-grained assessment of tool utilisation capabilities.
% These evaluation methodologies offer insights into the effectiveness of LLMs in real-world scenarios, highlighting areas for further improvement and refinement. 
However, there exists a gap in the literature concerning the exploration of evaluation stability when evaluating the tool usage capabilities of LLMs.

% \paragraph{In-context Learning with LLMs.}
% % for response generation
% In-context learning has been a transformative technique enabling LLMs to perform a wide range of tasks without fine-tuning. \cite{radford2019language} introduced the GPT series, demonstrating the potential of transformer-based models for in-context learning across various natural language processing tasks. \cite{brown2020language} furthered this work with GPT-3, showcasing the ability of LLMs to perform tasks given a small number of examples, or even a single example, within the prompt. This capability forms the foundation of our approach to mimicking API behavior.

% \paragraph{Instruction Generation.}
% The evolution of LLMs has enabled not only the comprehension but also the generation of instructions, marking a significant leap in natural language understanding and generation. 
% Benchmarks and tasks designed around instruction generation, as proposed by \citet{wei2021finetuned} and \citet{mishra2021cross}, demonstrate the potential of LLMs to craft coherent and contextually relevant instructions. 
% Following works ~\citep{patil2023gorilla, qin2023toolllm, tang2023toolalpaca} directly prompt LLMs to generate high-quality instructions or user queries under tool learning scenarios, demonstrating the effectiveness of utilizing LLMs for real-world instruction generation.

% \paragraph{API Generation and Simulation.}

% \section{Discussion}
% The phenomenon may occur beyound tool learning
% The experiments can be put afterwards, Stability first

\section{Conclusion}
% In this paper, to improve the stability of ToolBench, we propose a new stable benchmark named StableToolBench. Starting by the analysis of ToolBench, we find that the original ToolBench suffers from instability of evaluation processes and API status, which leads to fluctuation of evaluated model performance. To tackle the problem, we firstly determined all solvable problems in advance. We then replace the real API server with a virtual API server, supported by GPT-4 simulation of API behaviours when real APIs are not available. Finally, we build a caching system to store the server responses to use for increased stability. Extensive experiments show that StableToolBench can provide much more stable model performance. The simulated APIs demonstrate considerable reality and diversity while the caching system plays an important role in increasing stability.

In this paper, we propose StableToolBench, a benchmark developed to enhance the stability of ToolBench. Our analysis identified instability issues in the evaluation processes of ToolBench and API status, causing variability in model performance assessments. To address this, we implement a caching system for consistent data availability. We also replace the real API server with an LLM-simulated virtual server for reliable API behaviour simulation. Experiments show that StableToolBench significantly improves the stability of model performance evaluations, with the simulated APIs offering realism and the caching system contributing greatly to the enhanced stability of the benchmark.

% \clearpage
\section*{Acknowledgement}
This work is supported by the National Natural Science Foundation of China (No. 62276152, 61925601).
We also extend our gratitude to Jingwen Wu and Yao Li for their assistance with human evaluation and additional suggestions.
 
\section*{Limitations}
In this work, we propose StableToolBench, a new tool learning benchmark with increased stability but non-declining reality. However, our work faces the following limitations. Firstly, we used GPT-4 as the automatic evaluator in the evaluation process and as the backbone server, which increase the cost of using our benchmark. 
% Secondly, despite careful designs, including fixed version number and other parameters, GPT-4 as an backbone server still suffers from subtle instability when OpenAI updates its model. Though the impact of such instability is limited, we aim to further improve it by training an open-source model in the future. 
Secondly, the GPT-4 backboned server demonstrate strong performance in simulating API behaviours. Nevertheless, the backbone LLM may experience significant upgrades, which may affect the performance.
Therefore, we believe that the ultimate solution is to solve the problem with a trained open-source LLM. However, current open-source LLMs are not performant enough to simulate API behaviours well. As a result, closed-source LLMs are the only options. 
We believe that when open-source LLMs are strong enough to be well suited for this task, 
In the future, we may turn to open-source LLMs when they are strong enough to be well suited for this task.
Thirdly, although the cache hit rates are high with our explored methods, new methods will be developed in the future. Whether this cache will still be effective is unsure. In this regard, we aim to continuously update the cache in the future in a slow pace for both balanced stability and effectiveness.

% Nevertheless, the backbone LLM may experience significant upgrades, which may affect the performance.
% Therefore, we believe that the ultimate solution is to solve the problem with a trained open-source LLM. However, current open-source LLMs are not performant enough to simulate API behaviours well. As a result, closed-source LLMs are the only options. 
% % We believe that when open-source LLMs are strong enough to be well suited for this task, 
% In the future, we may turn to open-source LLMs when they are strong enough to be well suited for this task.

\bibliography{custom}

\begin{thebibliography}{37}
\expandafter\ifx\csname natexlab\endcsname\relax\def\natexlab#1{#1}\fi

\bibitem[{Brown et~al.(2020)Brown, Mann, Ryder, Subbiah, Kaplan, Dhariwal, Neelakantan, Shyam, Sastry, Askell, Agarwal, Herbert{-}Voss, Krueger, Henighan, Child, Ramesh, Ziegler, Wu, Winter, Hesse, Chen, Sigler, Litwin, Gray, Chess, Clark, Berner, McCandlish, Radford, Sutskever, and Amodei}]{brown2020language}
Tom~B. Brown, Benjamin Mann, Nick Ryder, Melanie Subbiah, Jared Kaplan, Prafulla Dhariwal, Arvind Neelakantan, Pranav Shyam, Girish Sastry, Amanda Askell, Sandhini Agarwal, Ariel Herbert{-}Voss, Gretchen Krueger, Tom Henighan, Rewon Child, Aditya Ramesh, Daniel~M. Ziegler, Jeffrey Wu, Clemens Winter, Christopher Hesse, Mark Chen, Eric Sigler, Mateusz Litwin, Scott Gray, Benjamin Chess, Jack Clark, Christopher Berner, Sam McCandlish, Alec Radford, Ilya Sutskever, and Dario Amodei. 2020.
\newblock \href {https://proceedings.neurips.cc/paper/2020/hash/1457c0d6bfcb4967418bfb8ac142f64a-Abstract.html} {{Language Models are Few-Shot Learners}}.
\newblock In \emph{Advances in Neural Information Processing Systems 33: Annual Conference on Neural Information Processing Systems 2020, NeurIPS 2020, December 6-12, 2020, virtual}.

\bibitem[{Cai et~al.(2024)Cai, Wang, Ma, Chen, and Zhou}]{cai2024large}
Tianle Cai, Xuezhi Wang, Tengyu Ma, Xinyun Chen, and Denny Zhou. 2024.
\newblock \href {https://openreview.net/forum?id=qV83K9d5WB} {{Large Language Models as Tool Makers}}.
\newblock In \emph{{Proc. of The Twelfth International Conference on Learning Representations (ICLR 2024)}}.

\bibitem[{Chen et~al.(2023{\natexlab{a}})Chen, Shu, Shareghi, Collier, Narasimhan, and Yao}]{chen2023fireact}
Baian Chen, Chang Shu, Ehsan Shareghi, Nigel Collier, Karthik Narasimhan, and Shunyu Yao. 2023{\natexlab{a}}.
\newblock \href {https://arxiv.org/abs/2310.05915} {{FireAct}: Toward language agent fine-tuning}.
\newblock \emph{ArXiv preprint}, abs/2310.05915.

\bibitem[{Chen et~al.(2023{\natexlab{b}})Chen, Du, Zhang, Liu, Liu, Zheng, Zhuo, Zhang, Lin, Chen et~al.}]{chen2023teval}
Zehui Chen, Weihua Du, Wenwei Zhang, Kuikun Liu, Jiangning Liu, Miao Zheng, Jingming Zhuo, Songyang Zhang, Dahua Lin, Kai Chen, et~al. 2023{\natexlab{b}}.
\newblock \href {https://arxiv.org/abs/2312.14033} {{T-Eval: Evaluating the Tool Utilization Capability Step by Step}}.
\newblock \emph{ArXiv preprint}, abs/2312.14033.

\bibitem[{{Gemini Team}(2023)}]{geminiteam2023gemini}
{Gemini Team}. 2023.
\newblock \href {http://arxiv.org/abs/2312.11805} {{Gemini: A Family of Highly Capable Multimodal Models}}.

\bibitem[{Gupta and Kembhavi(2023)}]{gupta2023visual}
Tanmay Gupta and Aniruddha Kembhavi. 2023.
\newblock Visual programming: Compositional visual reasoning without training.
\newblock In \emph{{In Proceedings of the IEEE/CVF Conference on Computer Vision and Pattern Recognition (CVPR 2023)}}, pages 14953--14962.

\bibitem[{Hao et~al.(2023)Hao, Liu, Wang, and Hu}]{hao2023toolkengpt}
Shibo Hao, Tianyang Liu, Zhen Wang, and Zhiting Hu. 2023.
\newblock \href {https://arxiv.org/abs/2305.11554} {{ToolkenGPT: Augmenting Frozen Language Models with Massive Tools via Tool Embeddings}}.
\newblock \emph{ArXiv preprint}, abs/2305.11554.

\bibitem[{Hsieh et~al.(2023)Hsieh, Chen, Li, Fujii, Ratner, Lee, Krishna, and Pfister}]{hsieh2023tool}
Cheng-Yu Hsieh, Si-An Chen, Chun-Liang Li, Yasuhisa Fujii, Alexander Ratner, Chen-Yu Lee, Ranjay Krishna, and Tomas Pfister. 2023.
\newblock \href {https://arxiv.org/abs/2308.00675} {Tool documentation enables zero-shot tool-usage with large language models}.
\newblock \emph{ArXiv preprint}, abs/2308.00675.

\bibitem[{Huang et~al.(2023)Huang, Shi, Li, Fan, Wu, Zhang, Liu, Zhou, Wan, Gong et~al.}]{huang2023metatool}
Yue Huang, Jiawen Shi, Yuan Li, Chenrui Fan, Siyuan Wu, Qihui Zhang, Yixin Liu, Pan Zhou, Yao Wan, Neil~Zhenqiang Gong, et~al. 2023.
\newblock \href {https://arxiv.org/abs/2310.03128} {{MetaTool} benchmark for large language models: Deciding whether to use tools and which to use}.
\newblock \emph{ArXiv preprint}, abs/2310.03128.

\bibitem[{Jin et~al.(2024)Jin, Yang, Chen, and Lu}]{jin2023genegpt}
Qiao Jin, Yifan Yang, Qingyu Chen, and Zhiyong Lu. 2024.
\newblock \href {https://doi.org/10.1093/bioinformatics/btae075} {{GeneGPT: augmenting large language models with domain tools for improved access to biomedical information}}.
\newblock \emph{Bioinformatics}, 40(2):btae075.

\bibitem[{Li et~al.(2023{\natexlab{a}})Li, Song, Yu, Yu, Li, Huang, and Li}]{li2023apibank}
Minghao Li, Feifan Song, Bowen Yu, Haiyang Yu, Zhoujun Li, Fei Huang, and Yongbin Li. 2023{\natexlab{a}}.
\newblock \href {http://arxiv.org/abs/2304.08244} {{API-Bank: A Benchmark for Tool-Augmented LLMs}}.

\bibitem[{Li et~al.(2023{\natexlab{b}})Li, Song, Yu, Yu, Li, Huang, and Li}]{li2023api}
Minghao Li, Feifan Song, Bowen Yu, Haiyang Yu, Zhoujun Li, Fei Huang, and Yongbin Li. 2023{\natexlab{b}}.
\newblock \href {https://arxiv.org/abs/2304.08244} {{API-Bank: A Comprehensive Benchmark for Tool-Augmented LLMs}}.
\newblock \emph{ArXiv preprint}, abs/2304.08244.

\bibitem[{Lu et~al.(2023)Lu, Peng, Cheng, Galley, Chang, Wu, Zhu, and Gao}]{lu2023chameleon}
Pan Lu, Baolin Peng, Hao Cheng, Michel Galley, Kai-Wei Chang, Ying~Nian Wu, Song-Chun Zhu, and Jianfeng Gao. 2023.
\newblock \href {https://arxiv.org/abs/2304.09842} {{Chameleon: Plug-and-Play Compositional Reasoning with Large Language Models}}.
\newblock \emph{ArXiv preprint}, abs/2304.09842.

\bibitem[{McInnes et~al.(2018)McInnes, Healy, Saul, and Grossberger}]{mcinnes2018umap-software}
Leland McInnes, John Healy, Nathaniel Saul, and Lukas Grossberger. 2018.
\newblock {UMAP: Uniform Manifold Approximation and Projection}.
\newblock \emph{The Journal of Open Source Software}, 3(29):861.

\bibitem[{Mialon et~al.(2023)Mialon, Dess{\`\i}, Lomeli, Nalmpantis, Pasunuru, Raileanu, Rozi{\`e}re, Schick, Dwivedi-Yu, Celikyilmaz et~al.}]{mialon2023augmented}
Gr{\'e}goire Mialon, Roberto Dess{\`\i}, Maria Lomeli, Christoforos Nalmpantis, Ram Pasunuru, Roberta Raileanu, Baptiste Rozi{\`e}re, Timo Schick, Jane Dwivedi-Yu, Asli Celikyilmaz, et~al. 2023.
\newblock \href {https://arxiv.org/abs/2302.07842} {{Augmented Language Models: A Survey}}.
\newblock \emph{ArXiv preprint}, abs/2302.07842.

\bibitem[{Nakano et~al.(2022)Nakano, Hilton, Balaji, Wu, Ouyang, Kim, Hesse, Jain, Kosaraju, Saunders, Jiang, Cobbe, Eloundou, Krueger, Button, Knight, Chess, and Schulman}]{nakano2022webgpt}
Reiichiro Nakano, Jacob Hilton, Suchir Balaji, Jeff Wu, Long Ouyang, Christina Kim, Christopher Hesse, Shantanu Jain, Vineet Kosaraju, William Saunders, Xu~Jiang, Karl Cobbe, Tyna Eloundou, Gretchen Krueger, Kevin Button, Matthew Knight, Benjamin Chess, and John Schulman. 2022.
\newblock \href {http://arxiv.org/abs/2112.09332} {{WebGPT: Browser-assisted question-answering with human feedback}}.

\bibitem[{OpenAI(2023)}]{openai2023gpt4}
OpenAI. 2023.
\newblock \href {http://arxiv.org/abs/2303.08774} {{GPT-4 Technical Report}}.

\bibitem[{Patil et~al.(2023)Patil, Zhang, Wang, and Gonzalez}]{patil2023gorilla}
Shishir~G. Patil, Tianjun Zhang, Xin Wang, and Joseph~E. Gonzalez. 2023.
\newblock \href {https://arxiv.org/abs/2305.15334} {{Gorilla: Large Language Model Connected with Massive APIs}}.
\newblock \emph{ArXiv preprint}, abs/2305.15334.

\bibitem[{Qin et~al.(2023{\natexlab{a}})Qin, Cai, Jin, Yan, Liang, Zhu, Lin, Han, Ding, Wang et~al.}]{qin2023webcpm}
Yujia Qin, Zihan Cai, Dian Jin, Lan Yan, Shihao Liang, Kunlun Zhu, Yankai Lin, Xu~Han, Ning Ding, Huadong Wang, et~al. 2023{\natexlab{a}}.
\newblock \href {https://arxiv.org/abs/2305.06849} {{WebCPM: Interactive Web Search for Chinese Long-form Question Answering}}.
\newblock \emph{ArXiv preprint}, abs/2305.06849.

\bibitem[{Qin et~al.(2023{\natexlab{b}})Qin, Hu, Lin, Chen, Ding, Cui, Zeng, Huang, Xiao, Han et~al.}]{qin2023tool}
Yujia Qin, Shengding Hu, Yankai Lin, Weize Chen, Ning Ding, Ganqu Cui, Zheni Zeng, Yufei Huang, Chaojun Xiao, Chi Han, et~al. 2023{\natexlab{b}}.
\newblock \href {https://arxiv.org/abs/2304.08354} {Tool learning with foundation models}.
\newblock \emph{ArXiv preprint}, abs/2304.08354.

\bibitem[{Qin et~al.(2023{\natexlab{c}})Qin, Liang, Ye, Zhu, Yan, Lu, Lin, Cong, Tang, Qian, Zhao, Tian, Xie, Zhou, Gerstein, Li, Liu, and Sun}]{qin2023toolllm}
Yujia Qin, Shihao Liang, Yining Ye, Kunlun Zhu, Lan Yan, Yaxi Lu, Yankai Lin, Xin Cong, Xiangru Tang, Bill Qian, Sihan Zhao, Runchu Tian, Ruobing Xie, Jie Zhou, Mark Gerstein, Dahai Li, Zhiyuan Liu, and Maosong Sun. 2023{\natexlab{c}}.
\newblock \href {http://arxiv.org/abs/2307.16789} {{ToolLLM: Facilitating Large Language Models to Master 16000+ Real-world APIs}}.

\bibitem[{Reimers and Gurevych(2019)}]{reimers-2019-sentence-bert}
Nils Reimers and Iryna Gurevych. 2019.
\newblock \href {https://doi.org/10.18653/v1/D19-1410} {Sentence-{BERT}: Sentence embeddings using {S}iamese {BERT}-networks}.
\newblock In \emph{Proceedings of the 2019 Conference on Empirical Methods in Natural Language Processing and the 9th International Joint Conference on Natural Language Processing (EMNLP-IJCNLP)}, pages 3982--3992, Hong Kong, China. Association for Computational Linguistics.

\bibitem[{Ruan et~al.(2023)Ruan, Chen, Zhang, Xu, Bao, Du, Shi, Mao, Zeng, and Zhao}]{ruan2023tptu}
Jingqing Ruan, Yihong Chen, Bin Zhang, Zhiwei Xu, Tianpeng Bao, Guoqing Du, Shiwei Shi, Hangyu Mao, Xingyu Zeng, and Rui Zhao. 2023.
\newblock \href {https://arxiv.org/abs/2308.03427} {{TPTU: Task planning and tool usage of large language model-based AI agents}}.
\newblock \emph{ArXiv preprint}, abs/2308.03427.

\bibitem[{Schick et~al.(2023)Schick, Dwivedi-Yu, Dess{\`\i}, Raileanu, Lomeli, Zettlemoyer, Cancedda, and Scialom}]{schick2023toolformer}
Timo Schick, Jane Dwivedi-Yu, Roberto Dess{\`\i}, Roberta Raileanu, Maria Lomeli, Luke Zettlemoyer, Nicola Cancedda, and Thomas Scialom. 2023.
\newblock \href {https://arxiv.org/abs/2302.04761} {{Toolformer: Language Models Can Teach Themselves to Use Tools}}.
\newblock \emph{ArXiv preprint}, abs/2302.04761.

\bibitem[{Song et~al.(2023)Song, Xiong, Zhu, Li, Wang, Tian, and Li}]{song2023restgpt}
Yifan Song, Weimin Xiong, Dawei Zhu, Cheng Li, Ke~Wang, Ye~Tian, and Sujian Li. 2023.
\newblock \href {https://arxiv.org/abs/2306.06624} {{RestGPT: Connecting Large Language Models with Real-World Applications via RESTful APIs}}.
\newblock \emph{ArXiv preprint}, abs/2306.06624.

\bibitem[{Tang et~al.(2023)Tang, Deng, Lin, Han, Liang, and Sun}]{tang2023toolalpaca}
Qiaoyu Tang, Ziliang Deng, Hongyu Lin, Xianpei Han, Qiao Liang, and Le~Sun. 2023.
\newblock \href {http://arxiv.org/abs/2306.05301} {{ToolAlpaca: Generalized Tool Learning for Language Models with 3000 Simulated Cases}}.

\bibitem[{Touvron et~al.(2023)Touvron, Lavril, Izacard, Martinet, Lachaux, Lacroix, Rozière, Goyal, Hambro, Azhar, Rodriguez, Joulin, Grave, and Lample}]{touvron2023llama}
Hugo Touvron, Thibaut Lavril, Gautier Izacard, Xavier Martinet, Marie-Anne Lachaux, Timothée Lacroix, Baptiste Rozière, Naman Goyal, Eric Hambro, Faisal Azhar, Aurelien Rodriguez, Armand Joulin, Edouard Grave, and Guillaume Lample. 2023.
\newblock \href {http://arxiv.org/abs/2302.13971} {{LLaMA: Open and Efficient Foundation Language Models}}.

\bibitem[{Turing(2009)}]{Turing2009}
Alan~M. Turing. 2009.
\newblock \href {https://doi.org/10.1007/978-1-4020-6710-5_3} {\emph{{Computing Machinery and Intelligence}}}, pages 23--65. Springer Netherlands, Dordrecht.

\bibitem[{Wang et~al.(2023)Wang, Wang, Liu, Chen, Yuan, Peng, and Ji}]{wang2023mint}
Xingyao Wang, Zihan Wang, Jiateng Liu, Yangyi Chen, Lifan Yuan, Hao Peng, and Heng Ji. 2023.
\newblock \href {https://arxiv.org/abs/2309.10691} {{MINT}: Evaluating {LLMs} in multi-turn interaction with tools and language feedback}.
\newblock \emph{ArXiv preprint}, abs/2309.10691.

\bibitem[{Wei et~al.(2023)Wei, Wang, Schuurmans, Bosma, Ichter, Xia, Chi, Le, and Zhou}]{wei2023chainofthought}
Jason Wei, Xuezhi Wang, Dale Schuurmans, Maarten Bosma, Brian Ichter, Fei Xia, Ed~Chi, Quoc Le, and Denny Zhou. 2023.
\newblock \href {http://arxiv.org/abs/2201.11903} {{Chain-of-Thought Prompting Elicits Reasoning in Large Language Models}}.

\bibitem[{Xu et~al.(2023)Xu, Hong, Li, Hu, Chen, and Zhang}]{xu2023tool}
Qiantong Xu, Fenglu Hong, Bo~Li, Changran Hu, Zhengyu Chen, and Jian Zhang. 2023.
\newblock \href {http://arxiv.org/abs/2305.16504} {{On the Tool Manipulation Capability of Open-source Large Language Models}}.

\bibitem[{Yang et~al.(2023{\natexlab{a}})Yang, Chen, Li, Ding, and Wu}]{yang2023chatgpt}
Linyao Yang, Hongyang Chen, Zhao Li, Xiao Ding, and Xindong Wu. 2023{\natexlab{a}}.
\newblock \href {https://arxiv.org/abs/2306.11489} {{ChatGPT is not Enough: Enhancing Large Language Models with Knowledge Graphs for Fact-aware Language Modeling}}.
\newblock \emph{ArXiv preprint}, abs/2306.11489.

\bibitem[{Yang et~al.(2023{\natexlab{b}})Yang, Song, Li, Zhao, Ge, Li, and Shan}]{gpt4tools}
Rui Yang, Lin Song, Yanwei Li, Sijie Zhao, Yixiao Ge, Xiu Li, and Ying Shan. 2023{\natexlab{b}}.
\newblock \href {http://arxiv.org/abs/2305.18752} {{GPT4Tools: Teaching Large Language Model to Use Tools via Self-instruction}}.

\bibitem[{Yao et~al.(2022{\natexlab{a}})Yao, Chen, Yang, and Narasimhan}]{NEURIPS2022_82ad13ec_webshop}
Shunyu Yao, Howard Chen, John Yang, and Karthik Narasimhan. 2022{\natexlab{a}}.
\newblock \href {https://proceedings.neurips.cc/paper_files/paper/2022/file/82ad13ec01f9fe44c01cb91814fd7b8c-Paper-Conference.pdf} {{WebShop: Towards Scalable Real-World Web Interaction with Grounded Language Agents}}.
\newblock In \emph{{In Proceedings of the Advances in Neural Information Processing Systems (NeurIPS 2022)}}, volume~35, pages 20744--20757. Curran Associates, Inc.

\bibitem[{Yao et~al.(2022{\natexlab{b}})Yao, Zhao, Yu, Du, Shafran, Narasimhan, and Cao}]{yao2023react}
Shunyu Yao, Jeffrey Zhao, Dian Yu, Nan Du, Izhak Shafran, Karthik Narasimhan, and Yuan Cao. 2022{\natexlab{b}}.
\newblock \href {https://arxiv.org/abs/2210.03629} {{ReAct}: Synergizing reasoning and acting in language models}.
\newblock volume abs/2210.03629.

\bibitem[{Ye et~al.(2024)Ye, Li, Gao, Huang, Wu, Li, Fan, Dou, Zhang, Gui, and Huang}]{ye2024tooleyes}
Junjie Ye, Guanyu Li, Songyang Gao, Caishuang Huang, Yilong Wu, Sixian Li, Xiaoran Fan, Shihan Dou, Qi~Zhang, Tao Gui, and Xuanjing Huang. 2024.
\newblock \href {http://arxiv.org/abs/2401.00741} {{ToolEyes: Fine-Grained Evaluation for Tool Learning Capabilities of Large Language Models in Real-world Scenarios}}.

\bibitem[{Zhuang et~al.(2023)Zhuang, Yu, Wang, Sun, and Zhang}]{zhuang2023toolqa}
Yuchen Zhuang, Yue Yu, Kuan Wang, Haotian Sun, and Chao Zhang. 2023.
\newblock \href {https://arxiv.org/abs/2306.13304} {{ToolQA: A Dataset for LLM Question Answering with External Tools}}.
\newblock \emph{ArXiv preprint}, abs/2306.13304.

\end{thebibliography}

\clearpage
\appendix

\section{Comparison of Reported and Reproduced Performance}
Detailed comparison scores of reported and reproduced performance are shown in \Cref{tab:performance_comparison}.

\begin{table}[ht!]
    \centering
    \small
    \begin{tabular}{lcc}
        \toprule
        {\textbf{Method}} & Reported & Reproduced \\
        \midrule
         GPT 3.5 Turbo 0613 + CoT & 41.5 & 35.2 \textcolor{red}{{\tiny 
         -32.5\%}} \\
         GPT 3.5 Turbo 0613 + DFS & 54.5 & 53.2 \textcolor{red}{{\tiny 
         -2.4\%}}\\
         ToolLLaMA v2 + CoT & 25.0 & 15.0 \textcolor{red}{{\tiny 
         -40\%}} \\
         ToolLLaMA v2 + DFS & 57.0 & 34.0 \textcolor{red}{{\tiny 
         -40.4\%}}\\
         \bottomrule
    \end{tabular}
    \caption{Comparison of performance (Pass Rate) reported in the paper and reproduced by us of ChatGPT and ToolLLaMA v2 on the I1-Instruction group of ToolBench. }
    \label{tab:performance_comparison}
\end{table}

\section{Statistics of API change information}
Detailed statistics of API change categories and information are shown in \Cref{tab:api_change} and \Cref{tab:api_not_available}.

\begin{table}[h!]
    \centering
    \small
    \begin{tabular}{lcc}
     \toprule
    \textbf{Status Type} & \textbf{Number} & \textbf{Percentage} (\%) \\
    \midrule
    % Not Available & 7504 & 45.6 \\
    Not Available & 8095 & 49.2 \\
    % \quad-- Not Connectable & 2426 & 14.7 \\
    % \quad-- Not Found & 583 & 3.5 \\
    % \quad-- Parameter Issues & 591 & 3.6 \\
    % \quad-- Other issues & 4495 & 27.3 \\
    Not Authorised & 1058 & 6.4 \\
    % Parameter Change & 591 & 3.6 \\
    Success & 7311 & 44.4 \\
     \bottomrule
    \end{tabular}
    \caption{APIs changed in ToolBench.}
    \label{tab:api_change}
\end{table}

\begin{table}[h!]
    \centering
    \small
    \begin{tabular}{lcc}
     \toprule
    \textbf{Status Type} & \textbf{Number} & \textbf{Percentage} (\%) \\
    \midrule
    Not Connectable & 2426 & 30.0 \\
    Not Found & 583 & 7.2 \\
    Parameter Change & 591 & 7.3 \\
    Parsing Error & 4247 & 52.6 \\
    Other & 248 & 3.1 \\
    \midrule
    Total & 8095 & 100 \\
     \bottomrule
    \end{tabular}
    \caption{Categories of Not Availability in ToolBench.}
    \label{tab:api_not_available}
\end{table}

% \section{Stability Test Scores with Virtual API Systems}
% Detailed stability test scores are shown in \Cref{tab:real_api_stability_test}.

% \begin{table}[]
%     \centering
%     \small
%     \resizebox{\linewidth}{!}{
%     \begin{tabular}{lcccc}
%         \toprule
%         \multirow{2}{*}{\textbf{Method}} & \multicolumn{4}{c}{\textbf{Percentage of Failing Tools}}\\
%         % \cmidrule{2-5}
%           & 0\% & 10\% & 20\% & 50\% \\
%         \midrule
%          GPT 3.5 Turbo 0613 + CoT & 20.3{\tiny $\pm{0.8}$}  & 17.9{\tiny $\pm{1.2}$} & 16.3{\tiny $\pm{0.6}$} & 12.8{\tiny $\pm{1.7}$} \\
%          GPT 3.5 Turbo 0613 + DFS & 26.6{\tiny $\pm{0.3}$} & 23.9{\tiny $\pm{1.1}$} & 23.2{\tiny $\pm{1.0}$} & 16.3{\tiny $\pm{1.2}$}\\
%          GPT 4 0613 + CoT & 21.4{\tiny $\pm{0.5}$}  & 19.6{\tiny $\pm{0.9}$} & 15.5{\tiny $\pm{0.3}$} & 11.8{\tiny $\pm{0.8}$} \\
%          GPT 4 0613 + DFS &  24.2{\tiny $\pm{1.8}$}  & 24.0{\tiny $\pm{0.8}$} & 21.2{\tiny $\pm{1.8}$} & 16.9{\tiny $\pm{0.4}$}\\
%          \bottomrule
%     \end{tabular}
%     }
%     \caption{SoPR change when manually make APIs down on the I1 Instruction group.}
%     \label{tab:real_api_stability_test}
% \end{table}
\section{Stability Test Scores with Virtual API Systems}
\label{app:detailed_stability_test_virtual}
Detailed scores of stability tests of various models are shown in \Cref{tab:simulated_api_stability_test}. Note that in addition to GPT 3.5 Turbo 0613 and GPT 4 0613, we report the performance of newer versions, namely GPT 3.5 Turbo 1106 and GPT 4 Turbo Preview.
\begin{table}[]
    \centering
    \small
    \resizebox{\linewidth}{!}{
    \begin{tabular}{ccccc}
        \toprule
        \multirow{2}{*}{\textbf{Method}} & \multicolumn{4}{c}{\textbf{Real API Failure Rate}}\\
        \cmidrule{2-5}
          & 0\% & 10\% & 20\% & 50\% \\
        \midrule
         GPT 3.5 Turbo 0613 + CoT & 49.1{\tiny $\pm{1.0}$}  & 48.7{\tiny $\pm{0.9}$} & 51.2{\tiny $\pm{1.3}$} & 49.0{\tiny $\pm{0.7}$} \\
         GPT 3.5 Turbo 0613 + DFS & 68.1{\tiny $\pm{1.4}$}  & 70.9{\tiny $\pm{1.3}$} & 67.5{\tiny $\pm{1.8}$} & 67.3{\tiny $\pm{1.3}$}\\
         GPT 4 0613 + CoT & 55.4{\tiny $\pm{0.6}$}  & 55.5{\tiny $\pm{1.0}$} & 58.0{\tiny $\pm{0.5}$} & 55.2{\tiny $\pm{0.6}$} \\
         GPT 4 0613 + DFS & 69.7{\tiny $\pm{1.4}$}  & 71.4{\tiny $\pm{1.4}$} & 71.2{\tiny $\pm{0.9}$} & 69.9{\tiny $\pm{0.9}$}\\
         \midrule
         GPT 3.5 Turbo 1106 + CoT & 52.1{\tiny $\pm{0.7}$}  & 52.4{\tiny $\pm{0.8}$} & 53.9{\tiny $\pm{0.6}$} & 50.2{\tiny $\pm{0.6}$} \\
         GPT 3.5 Turbo 1106 + DFS & 69.9{\tiny $\pm{0.7}$}  & 71.7{\tiny $\pm{0.7}$} & 69.4{\tiny $\pm{0.8}$} & 71.6{\tiny $\pm{0.9}$}\\
         GPT 4 Turbo preview + CoT & 60.8{\tiny $\pm{0.7}$}  & 62.8{\tiny $\pm{0.5}$} & 64.2{\tiny $\pm{0.7}$} & 62.4{\tiny $\pm{0.5}$} \\
         GPT 4 Turbo preview + DFS& 73.2{\tiny $\pm{1.1}$}  & 76.7{\tiny $\pm{1.0}$} & 76.0{\tiny $\pm{0.8}$} & 74.2{\tiny $\pm{1.3}$}\\     
         \bottomrule
    \end{tabular}
    }
    \caption{Performance change when manually make APIs down with our virtual online API system. The results are averaged over all six groups. Solving rates are reported. We run each experiment one time and evaluate three times and take the average score.}
    \label{tab:simulated_api_stability_test}
\end{table}

\section{Call Error Identification and Cache Filtering Rule}\label{app:filter_rule}
We identify call errors and filter out invalid call to RapidAPI based on keyword occurences. In detail, we identify the following error:
\begin{itemize}
    \item Not Connected Error: when error information contains \texttt{HTTP} or the response infomation contains \texttt{HTTP error, connection, rate limit, time(d) out};
    \item Not Found Error: when the error information or response contains \texttt{not found, not available, API doesn't exists, Service Not Found, internal error} or 404 error message;
    \item Parameter Change: when the error information or response contains \texttt{parameter, parse, is not defined};
    \item Parsing Error: when the error information starts with \texttt{Function executing from};
    \item Not Authorised: when the error information or response contains \texttt{authoriz(s), unauthoriz(s), blocked user, unsubscribe, credential, disabled for your subscription, ACCESS\_DENIED} or 401, 403 error message;
    \item Other Errors: messages with non-empty error messages;
    \item Success: Other calls.
\end{itemize}
We consider all types of errors when identifying errors. However, when filtering the cache, we do not conside the``Other Errors''.

\begin{table*}[ht!]
    % \small
    \centering
    % \resizebox{\columnwidth}{!}{
    \begin{tabular}{p{0.1\textwidth}p{0.8\textwidth}}
    \toprule
    \rowcolor[gray]{0.95} 
    \multicolumn{2}{c}{\textbf{API Simulation Prompt}} \\
    \midrule
    System & \makecell[{{p{.8\textwidth}}}]{
    Imagine you are an API Server operating within a specialized tool, which contains a collection of distinct APIs. Your role is to deeply understand the function of each API based on their descriptions in the API documentation. As you receive specific inputs for individual API calls within this tool, analyze these inputs to determine their intended purpose. Your task is to craft a JSON formatted response that aligns with the expected output of the API, guided by the provided examples. \\
    Your responses must adhere to a specific JSON structure, which is as follows: \\
    \texttt{\{
        ``error'': ``'',
        ``response'': ``Your\_Response''
    \}}\\
The error field should remain empty, indicating no errors in processing. The response field should contain the content you formulate based on the API's functionality and the input provided. Ensure that your responses are meaningful, directly addressing the API's intended functionality. If the provided examples are mostly error messages or lack substantial content, use your judgment to create relevant and accurate responses. The key is to maintain the JSON format's integrity while ensuring that your response is an accurate reflection of the API's intended output within the tool.\\
Please note that your answer should not contain anything other than a json format object, which should be parsable directly to json. \\
Note that: \\
- your response should be around 100 to 200 words, containing rich information given the api input parameters. Keep Your answer short and simple.\\
- your response must be effective and have practical content.\\
- if the api response example if null or ineffective, ignore the example and give your independent response. \\
    } \\
    \hline
    User & \makecell[{{p{.85\linewidth}}}]{
    API Documentation:\\
    \texttt{Documentation JSON file}\\
    API Examples: \\
    \texttt{Example input 1: Example response 1}\\
    \texttt{Example input 2: Example response 2}\\
    \texttt{Example input 3: Example response 3}\\
    API input:\\
    \texttt{Argument JSON string, e.g:} \\
    \texttt{\{``category'':``Logistics'',}\texttt{``tool\_name'': ``SQUAKE'',}\\
    \texttt{``api\_name'': ``Checkhealth'',``tool\_input'': ``\{\}'',}\\
    \texttt{``strip'': ``filter''\}}
    } \\
    \bottomrule
    \end{tabular}
    \caption{Prompt used to simulate APIs.}
    \label{tab:prompt_simulate_api}
\end{table*}

\begin{table*}[ht!]
    % \small
    \centering
    % \resizebox{\columnwidth}{!}{
    \begin{tabular}{l}
    \toprule
    \rowcolor[gray]{0.95} 
    \textbf{Solvable Task Filtration Prompt} \\
    \midrule
    \makecell[l{p{\textwidth}}]{
    Please check whether the given task solvable with following rules:\\
    1. If the \texttt{query} provide invalid information (e.g. invalid email address or phone number), return \texttt{Unsolvable}\\
    2. If the \texttt{query} needs more information to solve (e.g. the target restaurant name in a navigation task), return \texttt{Unsolvable} \\
    3. If the current \texttt{available\_tools} are enough to solve the query, return \texttt{Solvable} \\
    4. Return only \texttt{Solvable} or \texttt{Unsolvable} \\
    \\
    Task:\{\texttt{task}\}
    \\
    Now please give your answer (only \texttt{Solvable} or \texttt{Unsolvable}):
}
    \\
    \bottomrule
    \end{tabular}
    \caption{Prompt used to filter solvable tasks.}
    \label{tab:task_solvability}
\end{table*}

\section{Configurations of API Diversity Analysis }\label{app:diversity_conf}
The configurations of diversity analysis are as follows:
\begin{itemize}
    \item Embedding model: \texttt{all-mpnet-base-v2};
    \item UMAP metric (distance metric): correlation;
    \item Num of neighbours: 15;
    \item Min distance: 0.5.
\end{itemize}

\section{Prompts of API simulation}
\label{app:prompt_simulation}

The prompt used to simulate API behaviours is shown in \Cref{tab:prompt_simulate_api}.

\section{Prompt to Filter Solvable Task}
\label{app:prompt_task_solvability}
The prompt used to filter solvable tasks is shown in \Cref{tab:task_solvability}.

\section{Prompt Used to Make API Calls}\label{app:prompt_make_call}
The prompt used to construct API calls to scan availables is shown in \Cref{tab:prompt_api_call}.

\begin{table*}[t!]
    % \small
    \centering
    % \resizebox{\columnwidth}{!}{
    \begin{tabular}{p{0.1\textwidth}p{0.8\textwidth}}
    \toprule
    \rowcolor[gray]{0.95} 
    \multicolumn{2}{c}{\textbf{API Call Writing Prompt}} \\
    \midrule
    System & \makecell[{{p{.8\textwidth}}}]{
Imagine you are an API requester, Your role is to deeply understand the function of each API based on their descriptions in the API documentation.  Your task is to craft a JSON formatted input that aligns with the expected input of the API, guided by the provided examples.\\
Your responses must adhere to a specific JSON structure, which is as follows:\\
Please note that your answer should not contain anything other than a json format object, which should be parsable directly to json. \\
Note that:\\
- your response should be around 100 to 500 words, containing rich information given the api input parameters.\\
- your response must be effective and have practical content.\\
- if the api response example if null or ineffective, ignore the example and give your independent response.\\
    } \\
    \hline
    User & \makecell[{{p{.85\linewidth}}}]{
    API Documentation:\\
    \texttt{Documentation JSON file}\\
    API Examples (if available): \\
    \texttt{Example input 1: Example response 1}\\
    \texttt{Example input 2: Example response 2}\\
    \texttt{Example input 3: Example response 3}\\
    one more API Input example:\\
    } \\
    \bottomrule
    \end{tabular}
    \caption{Prompt used to write API calls.}
    \label{tab:prompt_api_call}
\end{table*}

% \section{Example Appendix}
% \label{sec:appendix}

% This is an appendix.

\end{document}